\pdfoutput=1

\documentclass[11pt]{article}

\usepackage{acl}
\usepackage{hyperref}

\usepackage{times}
\usepackage{latexsym}

\usepackage[T1]{fontenc}

\usepackage[utf8]{inputenc}
\usepackage{url}
\usepackage{graphicx}
\usepackage{booktabs}
\usepackage{subcaption}
\usepackage{enumitem}
\usepackage{listings}

\usepackage{tabularx,arydshln}
\usepackage{color,soul}

\usepackage{microtype}

\title{What Really Counts? Examining Step and Token Level Attribution in Multilingual CoT Reasoning}

\author{Jeremias Ferrao ~\;~ Ezgi Basar ~\;~ Khondoker Ittehadul Islam ~\;~ Mahrokh Hassani\\[5pt]University of Groningen\\[5pt]
  {\tt \{j.l.ferrao, e.basar, k.i.islam, m.hassani\}}{\tt@student.rug.nl}\\
}

\begin{document}
\maketitle
\begin{abstract}
This study investigates the attribution patterns underlying Chain-of-Thought (CoT) reasoning in multilingual LLMs. While prior works demonstrate the role of CoT prompting in improving task performance, there are concerns regarding the faithfulness and interpretability of the generated reasoning chains. To assess these properties across languages, we applied two complementary attribution methods—ContextCite for step-level attribution and Inseq for token-level attribution—to the Qwen2.5 1.5B-Instruct model using the MGSM benchmark. Our experimental results highlight key findings such as: (1) attribution scores excessively emphasize the final reasoning step, particularly in incorrect generations; (2) structured CoT prompting significantly improves accuracy primarily for high-resource Latin-script languages; and (3) controlled perturbations via negation and distractor sentences reduce model accuracy and attribution coherence. These findings highlight the limitations of CoT prompting, particularly in terms of multilingual robustness and interpretive transparency. To facilitate reproducibility, we make our code available at \url{https://github.com/Jazhyc/IKNLP-Attribution}.
\end{abstract}




\section{Introduction}

Large Language Models (LLMs) have demonstrated remarkable versatility across various natural language tasks \cite{gpt2}. A significant advancement in enhancing their complex reasoning abilities is Chain-of-Thought (CoT) prompting \cite{wei2022chain}, which guides models to generate intermediate reasoning steps, often leading to substantial performance improvements \cite{livebench, wang2022self}. Recent studies show that the generated reasoning chains are often inconsistent and prone to producing misleading intermediate steps, thereby casting doubt on their explanatory reliability \cite{lanham2023measuringfaithfulnesschainofthoughtreasoning}.




Feature attribution techniques break down a model's internal processes by assigning importance scores on various segments, such as reasoning steps, to understand their contribution to the final answer \cite{attribution-survey, rashkin2023measuring, bohnet2022attributed}. Despite their promise, attribution techniques have seldom been applied to assess reasoning fidelity in CoT across languages and at different levels of granularity (step-wise vs. token-level) \cite{cot-attribution}.

In this paper, we investigate the consistency and interpretability of multilingual LLM reasoning under CoT prompting. We leverage both step-wise and token-level attribution to uncover model vulnerabilities and strategy variations across linguistic contexts.

Our research objectives are summarized as follows: (1) analyzing and comparing the importance of reasoning steps across languages, (2) investigating how attribution patterns differ between correct and incorrect predictions to understand variations in reasoning strategies, and (3) performing token-level attribution to examine how controlled modifications to the input affect the reasoning process.

\section{Related Work}






\paragraph{LLM and CoT}
LLMs operate as probabilistic sequence predictors, estimating the likelihood of the next token given previous context. While grounded in principles from information theory \cite{shannon1948mathematical,shannon1951prediction}, modern LLMs acquire linguistic and conceptual representations through massive-scale training. Despite their success, base LLMs often produce vague or inconsistent responses \cite{touvron2023llama}.

Instruction tuning further improves LLM behavior. A recent study, \citet{wang2022self}, showed that training with instruction–prompt–response triples enhances task fidelity and user understanding. Furthermore, \citet{wei2022chain} found that prompting instruction-based models to elaborate on their reasoning process—also, CoT reasoning—before delivering a final answer leads to improved accuracy in the GSM8K dataset. Finally, \citet{lanham2023measuringfaithfulnesschainofthoughtreasoning} examined whether CoT reasoning aligns with the predictions made by LLMs. They tested robustness to perturbations by adding mistakes in the reasoning steps. They found that faithfulness to CoT depends on task complexity and model scale, with larger models relying less on their own reasoning.

\paragraph{Cross-Linguality}


Performance disparities in multilingual LLMs stem from uneven language representation in pretraining data. Well-resourced languages using Latin scripts (e.g., English, French, German) consistently outperform underrepresented or low-resource languages such as Chinese and Bengali \cite{chau2021specializing, ahia2023all, nguyen2024prompting, ahia2024magnet}. These discrepancies are further amplified by script-related tokenization inefficiencies, which can hinder downstream task performance \cite{ahia2023all}.

\paragraph{Attribution}

Recently, \citet{hu2023reveal} demonstrated that specific reasoning steps in CoT are more predictive of final answer correctness. While their findings pertain to natural language outputs, we investigated whether similar patterns arise in numerical reasoning. Additionally, \citet{cobbe2021training} highlighted the importance of token-level attribution in generating final answers. Since we worked with a similar dataset, we further investigated trends in token significance.











\section{Method}
\label{sec:method}

Our approach combines language model prompting techniques with post-hoc attribution analysis to investigate multilingual CoT reasoning in mathematical word problems.

\subsection{Language Model}

We selected Qwen2.5-1.5B-Instruct \cite{qwen2.5} as the language model for our experiments due to 1) Its small size (1.5 billion parameters), aligning with our computational resources, 2) its multilingual support, vital for assessing CoT reasoning across diverse languages, and 3) its impressive benchmark scores compared to other models of similar parameter counts \cite{qwen2.5}. We specifically used the instruction-tuned variant in the main report due to its adaptability on complex instructions like CoT prompting \cite{wang2022self}. However, we also report results of the DeepSeek distilled variant of the model in Appendix \ref{sec:deepseek}.

\subsection{Step-Wise Attribution using ContextCite}
To analyze the contribution of each generated reasoning step to the final answer, we employed ContextCite \cite{contextcite} due to its flexibility in operating directly at the level of user-defined text segments, such as the sentences constituting our CoT steps. This contrasts with token-level attribution methods like Integrated Gradients \cite{integrated-gradients} or attention map analysis \cite{attention-attribution}, which first assign scores to individual tokens that must then be aggregated (e.g., summed or averaged) to estimate sentence-level importance, potentially losing step-specific signal. Furthermore, ContextCite is a post-hoc method that operates directly on the pre-trained LLM and its generated output. It does not require modification of the model architecture or retraining \cite{rashkin2023measuring, bohnet2022attributed}, making it readily applicable to existing models and responses.

ContextCite determines step importance by treating the generated reasoning steps as ``context'' segments and systematically ablating various combinations of them. For each ablation, it measures the language model’s probability of generating the original final answer given the remaining steps. It then fits a sparse linear surrogate model using LASSO regression to predict this probability based on the presence or absence of each step. The coefficients of this surrogate model serve as attribution scores, directly quantifying the contributive importance of each reasoning step to the model’s conclusion.

\subsection{Token-Level Attribution using Inseq}

Our token-level attribution experiments use the saliency feature attribution method \cite{saliency-map} implemented in the Inseq toolkit \citep{Sarti_2023}. This technique measures how sensitive the model's predictions are to each source and target token. Using this method, the attribution scores are generated by computing the gradient of the model's predicted probability for a given output token concerning the input embeddings.

For our analysis, we need to observe how certain sequences affect a generated token. The toolkit handles this through its aggregation function. By default, this function sums attribution scores along the final dimension and normalizes these values by dividing them by the norm of the source and target attributions. This gives us a meaningful way to identify which sentences or phrases are more influential when generating the answer rather than simply looking at the relationship between individual tokens.

\section{Experimental Setup}

This section details the methodology employed to investigate CoT reasoning across multiple languages and analyze the contribution of reasoning steps using attribution techniques.

\subsection{Dataset and Preprocessing}
We utilized the Multilingual Grade School Math (MGSM) dataset \cite{mgsm}, an extension of the English-only GSM8K benchmark \cite{gsm8k}. GSM8K consists of grade-school level math word problems that are typically solvable by middle school students in 2 to 8 reasoning steps. MGSM contains 250 such problems from the GSM8K test set, manually translated into ten other typologically diverse languages. Due to time and computational constraints, we limited our experiments to five languages: English (EN), French (FR), German (DE), Bengali (BN), and Chinese (ZH). We selected these languages because (1) we were already familiar with English, French, German, and Bengali; (2) the Qwen models were extensively tested on Chinese benchmarks \cite{qwen2.5}; and (3) we aimed to include languages with diverse scripts for a broader comparison.

Following the CoT practices of \citet{mgsm}, we employed few-shot in-context learning. For each target language, we selected all 8 examples from the MGSM training split to promote CoT reasoning. Each of these examples adheres to a specific format: they begin with a language-specific preamble (e.g., "Step-by-Step Answer:" in English), followed by the intermediate reasoning steps, then a consistent prefix indicating the final answer (e.g., "The answer is"), and the numerical answer itself. To prepare these examples, we preprocessed the reasoning steps within them so that each step appeared on a new line, simplifying the use of structured generation (Section \ref{sec:structured-generation}). Our evaluations were subsequently performed on the MGSM test set, which contains 250 distinct questions per language.

\subsection{Language Model and Generation}


Generation was performed using the model's default parameters without specific tuning, obtaining answers in a single pass (``one-shot") for each test question due to resource limitations. We additionally enforce a token limit of 256 for the generated responses to ensure the quick execution of our experiments.

\subsection{Structured Generation}
\label{sec:structured-generation}
We used structured generation \cite{outlines} to enforce adherence to the desired CoT structure (Figure \ref{fig:mgsm-cot-structure}) and enable separation of reasoning steps. This technique allowed us to guide the model's output by restricting the next tokens to those that conform to a predefined regular expression. More information regarding how we use structured generation can be found in Appendix \ref{sec:sg-regex}.

\subsection{Evaluation Metrics and Baselines}
\label{sec:evaluation-contextcite}
Our primary evaluation metric is accuracy, calculated following \citet{mgsm}. The final numerical answer is extracted from the model's output and compared against the ground truth from the MGSM test set. A preliminary analysis on the English dataset was conducted to validate our approach and establish the impact of CoT and structured generation. We compared four setups:
\begin{enumerate}[leftmargin=*,align=left]
\item \textsc{\textbf{NoCoT-Unstruct:}} Model input is only the question; generation is unconstrained. This setup is identical to regular prompting.
\item \textsc{\textbf{CoT-Unstruct:}} Input includes 8 few-shot CoT examples and the question; generation is unconstrained.
\item \textsc{\textbf{NoCoT-Struct:}} Input is only the question; output is forced via structured generation to be only the final numerical answer. This baseline was specifically designed to account for the possibility that instruction-tuned models, such as those in the Qwen family \cite{qwen2.5}, might perform implicit CoT reasoning even without explicit prompting. Preliminary tests showing verbose baseline answers supported this concern. By constraining the output to only the final number, we aim to measure the model's performance based purely on its initial understanding, preventing contamination from inherently generated reasoning steps.
\item \textsc{\textbf{CoT-Struct:}} Input includes 8 few-shot CoT examples and the question; output is forced via structured generation to follow the CoT format (Figure \ref{fig:mgsm-cot-structure}).
\end{enumerate}


\begin{figure}[h]
    \centering
    \includegraphics[width=0.95\linewidth]{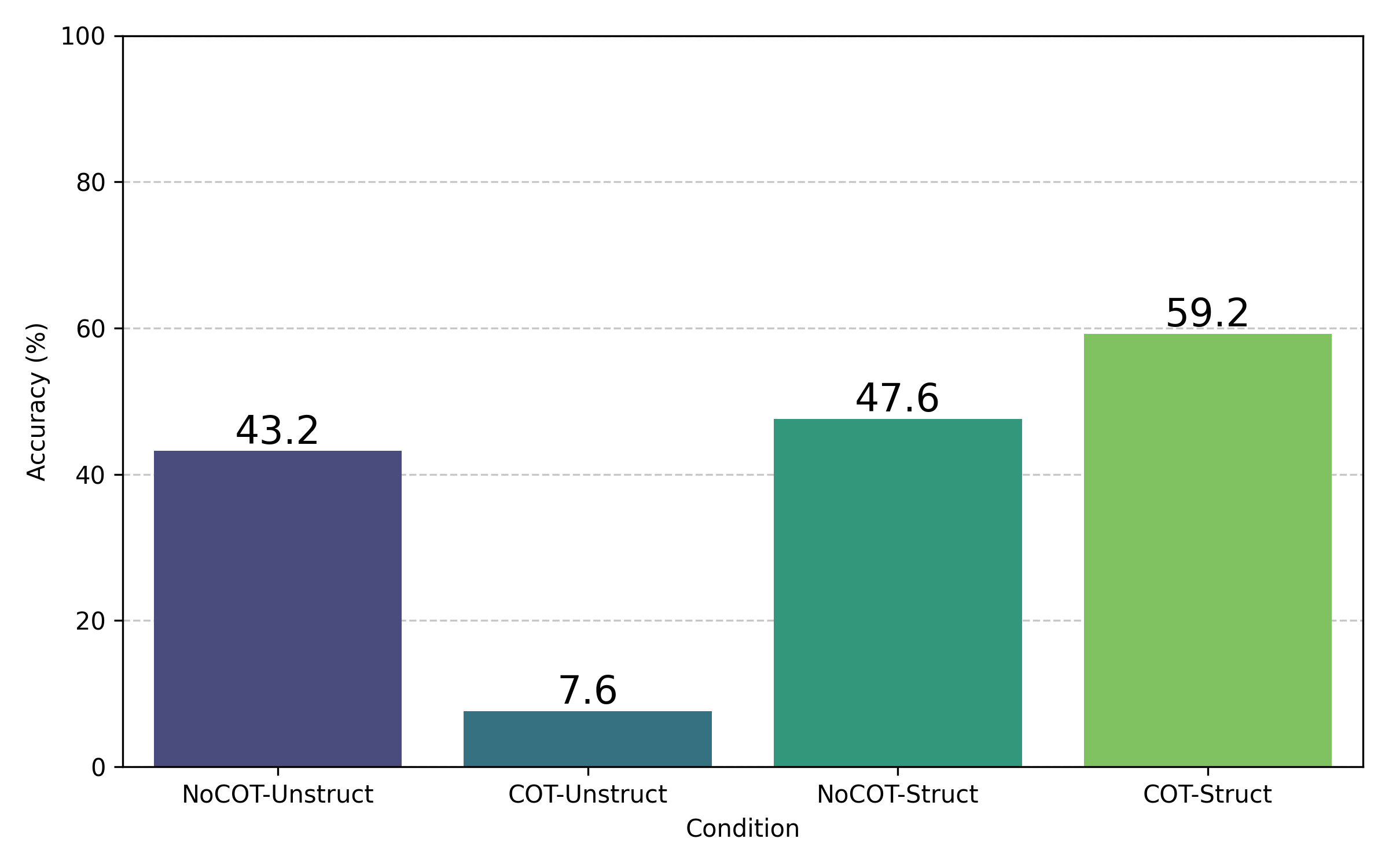}
    \caption{Preliminary evaluation results on the English portion of the MGSM dataset using the conditions described in Section \ref{sec:evaluation-contextcite}.}
    \label{fig:baseline}
\end{figure}

After conducting a preliminary analysis in English (Figure \ref{fig:baseline}), we observed significantly higher accuracy of \textsc{CoT-Struct} compared to all other setups. As a result, for the rest of our experiment, we primarily conduct experiments on the \textsc{CoT-Struct} as this allows us to isolate and assess the contribution of each reasoning step to different languages vital to fulfilling our research objective. Additionally, for baseline comparison, we consider \textsc{NoCoT-Unstruct} as this condition allows us to measure the aptitude of the model at obtaining the final answer without explicit reasoning.


\subsection{Step-Wise Attribution}
After generating the structured CoT responses across our tested languages, we applied ContextCite \cite{contextcite} to quantify the importance of each reasoning step towards the final generated answer. We configured the method to use 32 context ablations (instead of the default 64) per generated response to balance computational cost and attribution quality, following recommendations by \citet{contextcite}.

\section{Results \& Analysis}

This section presents the results of our experiments investigating CoT reasoning in the Qwen Instruct model across multiple languages using the MGSM dataset. We first characterize the model's generation behavior (accuracy, length) across languages, then delve into step-wise attribution patterns using ContextCite, and finally examine token-level attributions under different conditions using Inseq.

\subsection{Overall Results}

\begin{figure}[h]
    \centering
    \includegraphics[width=0.9\linewidth]{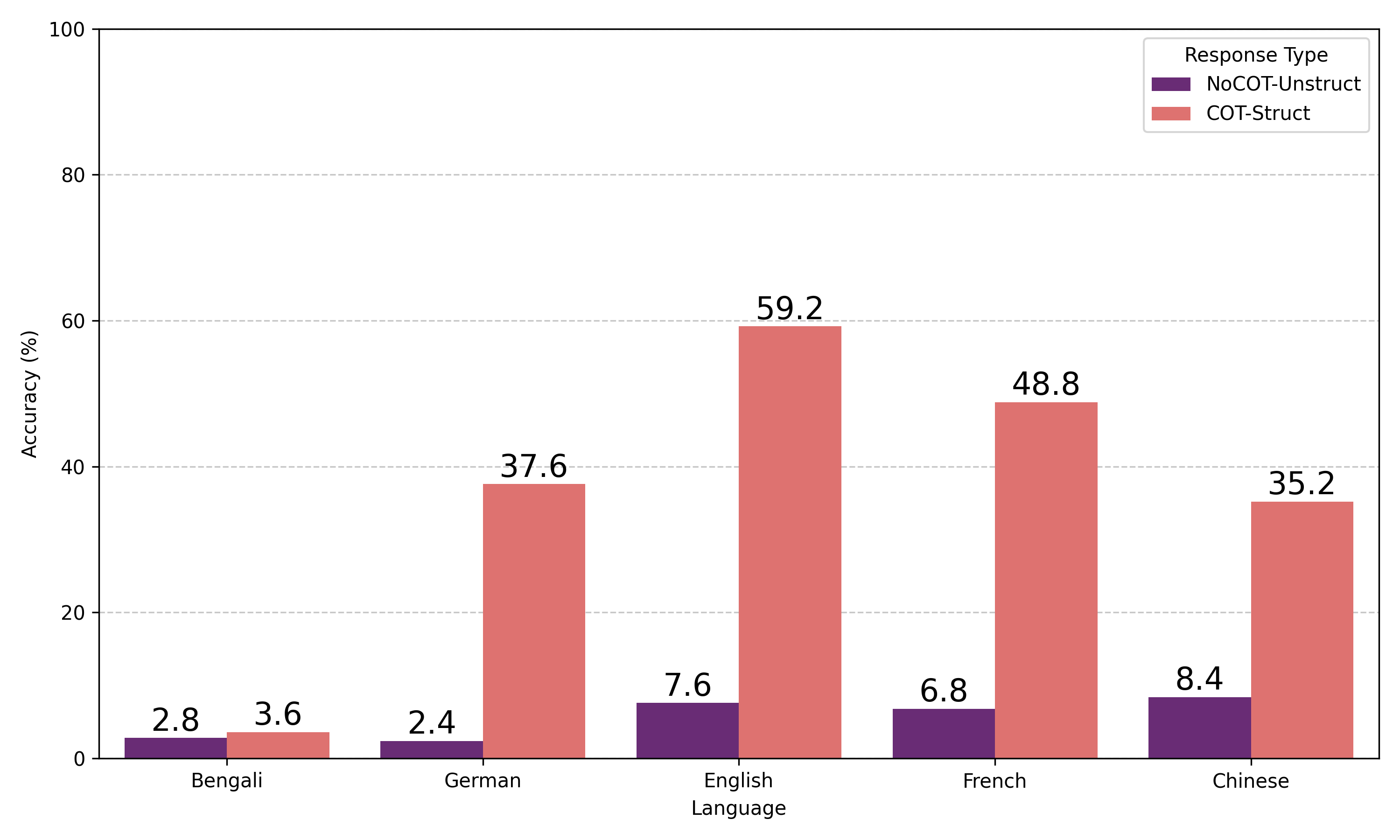}
    \caption{Accuracy results for Qwen Instruct on MGSM across five languages. Generating only a direct answer (NoCoT-Unstruct) results in low accuracy (<10\%). Introducing structured CoT (CoT-Struct) dramatically boosts performance, most significantly for English. However, this improvement trend is minimal for Bengali.}
    \label{fig:multilang-accuracy}
\end{figure}

We report the accuracy of our experimental results in Figure \ref{fig:multilang-accuracy}. All languages show improved performance on \textsc{CoT-Struct} compared to the baseline \textsc{NoCoT-Unstruct}. This aligns with our earlier analysis conducted only on English during the setup selection (see Figure \ref{fig:baseline}). Moreover, among all languages, English achieves the highest accuracy of $59.2$\% followed by French ($48.8$\%), German ($37.6$\%), and Chinese ($35.2$\%) respectively. Additionally, Bengali achieved the lowest accuracy ($3.6$\%), achieving only $\approx$0.8\% improvement over the baseline.

\begin{figure*}[htbp]
    \centering
    \begin{subfigure}[b]{\columnwidth}
        \centering
        \includegraphics[width=0.9\textwidth]{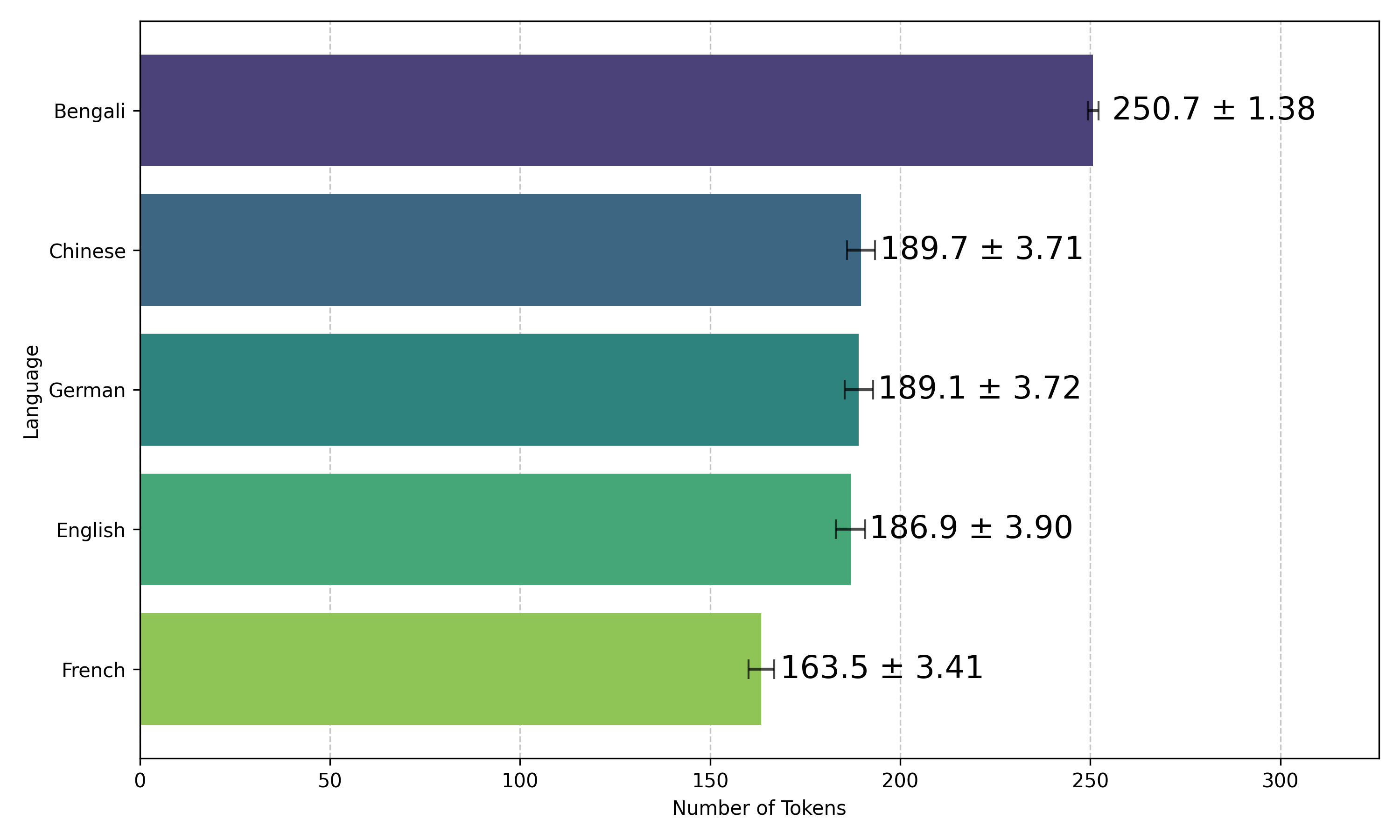}
        \label{fig:token-variation-mean}
    \end{subfigure}
    \hfill
    \begin{subfigure}[b]{\columnwidth}
        \centering
        \includegraphics[width=0.9\textwidth]{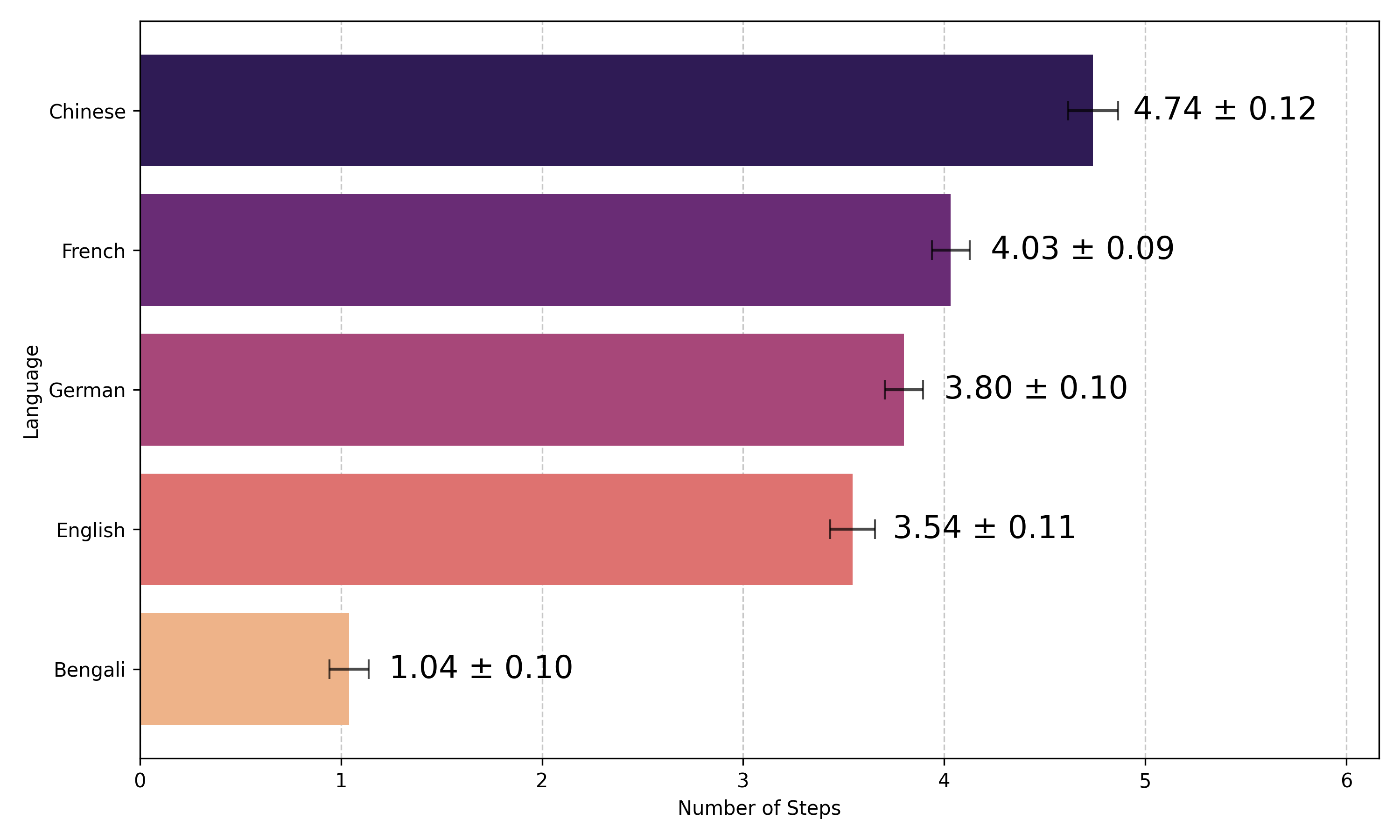}
        \label{fig:token-variation-std}
    \end{subfigure}
    \caption{Mean token count (left) and reasoning steps (right) produced by Model along with standard errors.}
    \label{fig:token-variation-nondistill}
\end{figure*}

\paragraph{Per Language Generation Length}

To analyze the performance trend of \textsc{CoT-Struct}, we examine the average token count and number of reasoning steps shown in Figure \ref{fig:token-variation-nondistill}. Notably, \textsc{CoT-Struct} in Bengali exhibits the highest average token count (250.7), slightly higher than Chinese (189.7). Conversely, the average number of reasoning steps in Bengali is much lower (1.04). This discrepancy is expected as Bengali is a non-Latin-script language, requiring the model to tokenize words into more sub-words. Additionally, as a low-resource language, the limited vocabulary available for the model hinders its ability to predict the next token and generate valid reasoning steps, resulting in either failure to reach an answer or the production of incorrect ones.


\begin{table}[h]
    \centering
    \begin{tabular}{lc}
    \toprule
    Language & Parsed Entries Ratio \\
    \midrule
    English & 0.54 \\
    Bengali & 0.64 \\
    German & 0.74 \\
    French & 0.92 \\
    Chinese & 0.98 \\
    \bottomrule
    \end{tabular}
    \caption{Success rate of structured generation across languages, measured as the ratio of responses perfectly matching the required regular expression format.}
    \label{tab:regex_adherence}
\end{table}

\paragraph{Structured Generation Compliance}

A prerequisite for conducting our step-wise attribution analysis with ContextCite is the successful generation of responses that strictly adhere to the predefined CoT structure. We measured this compliance by calculating the ratio of perfectly parsed responses per language (Table \ref{tab:regex_adherence}). The results show significant variation: French (0.92) and Chinese (0.98) demonstrated high adherence, while German (0.74), Bengali (0.64), and English (0.54) had lower success rates. Potential reasons for non-compliance include insufficient allocated tokens for the full format, the model generating incoherent steps (possibly due to its smaller size), or deviating from the structure by embedding the final answer within the reasoning chain instead of on the designated line (see Appendix \ref{sec:sg-regex} for examples of these failure cases). Generations that failed to be parsed were ignored for the ContextCite analysis. This differing ability to follow structured prompts highlights language-specific variations and limitations in constrained generation.

\subsection{Step-Wise Attribution (ContextCite)}
\label{sec:results_contextcite}

\begin{figure}[ht]
    \centering
    \includegraphics[width=0.9\linewidth]{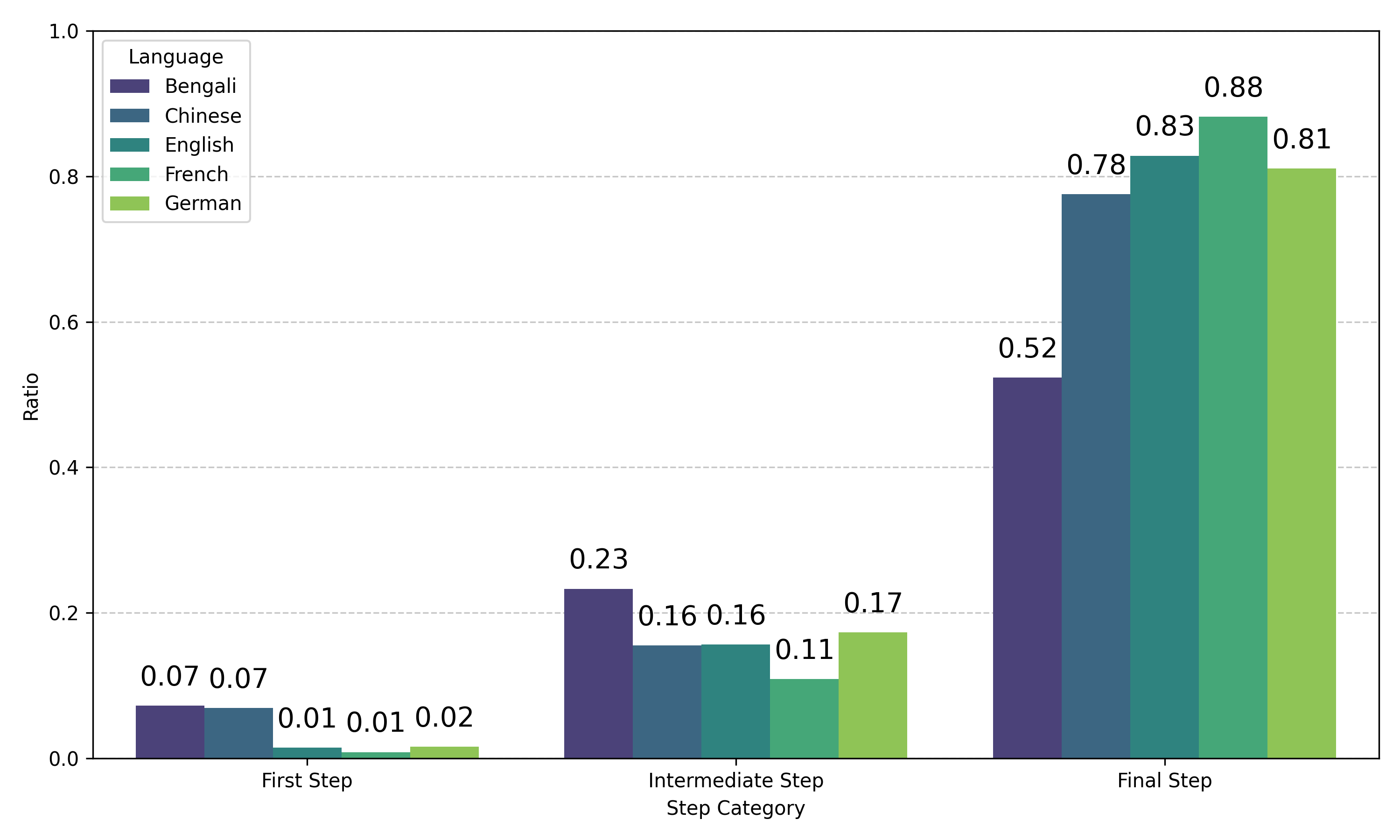}
    \caption{Distribution of the highest-attributed reasoning step category (First/Preamble, Intermediate, Final) for Qwen Instruct on MGSM, based on ContextCite scores across five languages.}
    \label{fig:step-importance-category}
\end{figure}

After establishing the performance and generation characteristics of our model, we apply ContextCite to analyze the importance of different reasoning steps. First, we examine which part of the reasoning chain typically receives the highest attribution score. Figure \ref{fig:step-importance-category} demonstrates the distribution of the highest-attributed step category (First/Preamble, Intermediate, Final). Across all five languages, the final reasoning step (the one immediately preceding the numerical answer) is most frequently identified as the most important contributor to the model's final answer prediction. This aligns with intuition, as this step usually involves the final calculation before reporting the answer.

\begin{table}[h]
    \centering
    \begin{tabular}{lcc}
    \toprule
    Language & Slope (Correct) & Slope (Incorrect) \\
    \midrule
    English & 5.24 & 6.54 \\
    Bengali & 9.31 & 12.73 \\
    German & 4.98 & 5.27 \\
    French & 4.59 & 4.47 \\
    Chinese & 4.96 & 4.44 \\
    \bottomrule
    \end{tabular}
    \caption{Slopes of linear regression models fitted to ContextCite step importance scores against normalized step position (0=First, 1=Last).}
    \label{tab:step_importance_slopes}
\end{table}

To gain a more granular understanding, we analyze the trend of step importance scores from the first step (normalized position 0) to the last step (normalized position 1). Figure \ref{fig:step-importance-langs} (Appendix \ref{sec:cc-importance}) visualizes these trends, plotting importance scores against normalized step position for both correct and incorrect predictions. We fitted linear regression models to these points to quantify the trend; the slopes are reported in Table \ref{tab:step_importance_slopes}.

For all languages, both correct and incorrect predictions generally show a positive slope, confirming that importance tends to increase linearly towards the final steps. Comparing correct versus incorrect predictions, we observe variations. For English, Bengali, and German, the slope is steeper for incorrect predictions (e.g., 12.73 for Bengali incorrect vs. 9.31 correct), suggesting the model may rely disproportionately on inaccurate final steps when it makes an error. In contrast, for French and Chinese, the slopes are very similar for correct and incorrect predictions, with the incorrect slope being slightly lower. Notably, Bengali exhibits the steepest slopes overall, especially for incorrect answers, while Chinese shows relatively shallow slopes, particularly for incorrect predictions (4.44).

\subsection{Token Level Attribution (Saliency)}

Our initial experiments showed that step importance increases as we move forward in the reasoning chain. To take a closer look at this pattern, we utilized the Inseq toolkit \citep{Sarti_2023} to get fine-grained attribution scores for the same Qwen model. We perform subsequent experiments on the MGSM training set with 8 items per language. Given the limited dataset size, we emphasize that these experiments only serve as a case study and the results should be carefully interpreted when extrapolating our observations to more representative datasets. We restricted our analysis to English, French, and German because these were the only languages that met both the model performance standards and our team’s working capacity. Unlike the previous approach, which employed few-shot prompting, we do not include any CoT examples in our prompt. This is due to the fact that we are working with the dataset that was previously used for few-shot prompts. 

The study performs controlled evaluations across three experimental conditions: (1) baseline CoT generation, (2) the negation condition, and (3) the distractor condition. The latter two are cases where the original question has been slightly modified to assess their impact on reasoning steps. This process was performed using dependency parsers from the spaCy library \citep{honnibal2020spacy} and then manually corrected. Similar to \citet{lanham2023measuringfaithfulnesschainofthoughtreasoning}, we investigate the impact of perturbations on CoT reasoning. However, while prior studies intervene directly on the CoT steps, we instead perturb the input question to analyze how these modifications alter attribution patterns in the generated CoT.

\begin{table}[ht]
\centering
\begin{small}
\begin{tabular}{@{}l p{5.3cm}@{}}
\toprule
\textbf{Condition} & \textbf{Question} \\
\midrule
Original & Roger has 5 tennis balls. He buys 2 more cans of tennis balls. Each can has 3 tennis balls. How many tennis balls does he have now? \\
Negation & Roger has 5 tennis balls. He \textit{does not} buy 2 more cans of tennis balls. Each can has 3 tennis balls. How many tennis balls does he have now? \\
Distractor & Roger has 5 tennis balls. He buys 2 more cans of tennis balls. Each can has 3 tennis balls. \textit{Roger drinks 3 cans of soda.} How many tennis balls does he have now? \\
\bottomrule
\end{tabular}
\end{small}
\caption{Examples of original, negation, and distractor versions of an English question from the MGSM dataset. Modifications are shown in italics.}
\label{tab:q_conditions}
\end{table}

For the negation condition, we select a sentence in the middle of the question and negate its main verb while maintaining subject agreement and tense markers. For the distractor condition, we insert an irrelevant sentence at the penultimate position of the question. This sentence is constructed using the first subject in the question. The distractor sentence, likewise, respects subject agreement and maintains tense consistency with the rest of the question. Examples of each question are provided in Table \ref{tab:q_conditions}.

\begin{figure}[htbp]
    \centering
    \begin{minipage}[b]{0.46\textwidth}
        \includegraphics[width=\textwidth]{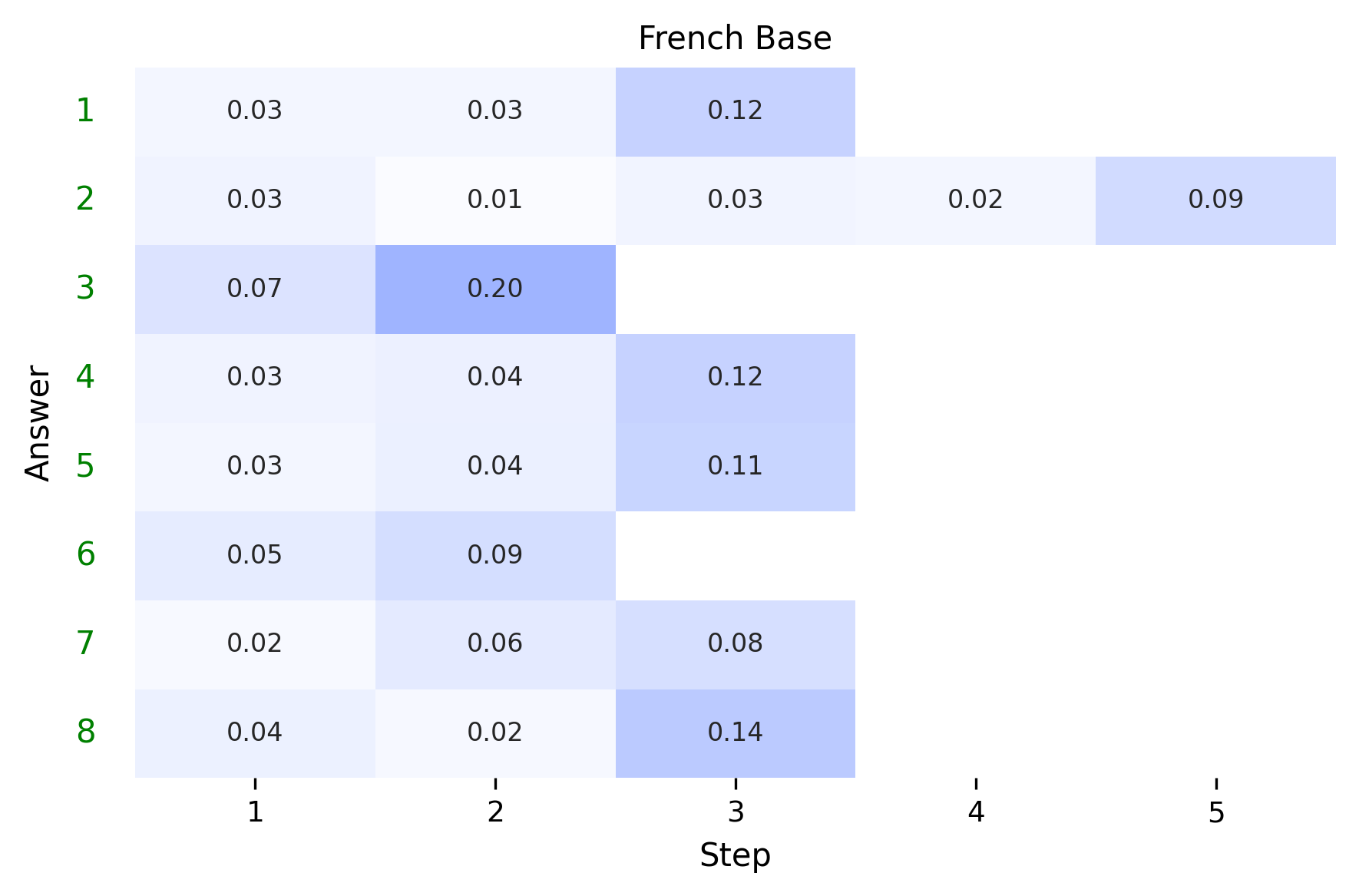}
    \end{minipage}
    
    \vspace{0.5em}
    \begin{minipage}[b]{0.46\textwidth}
        \centering
        \includegraphics[width=\textwidth]{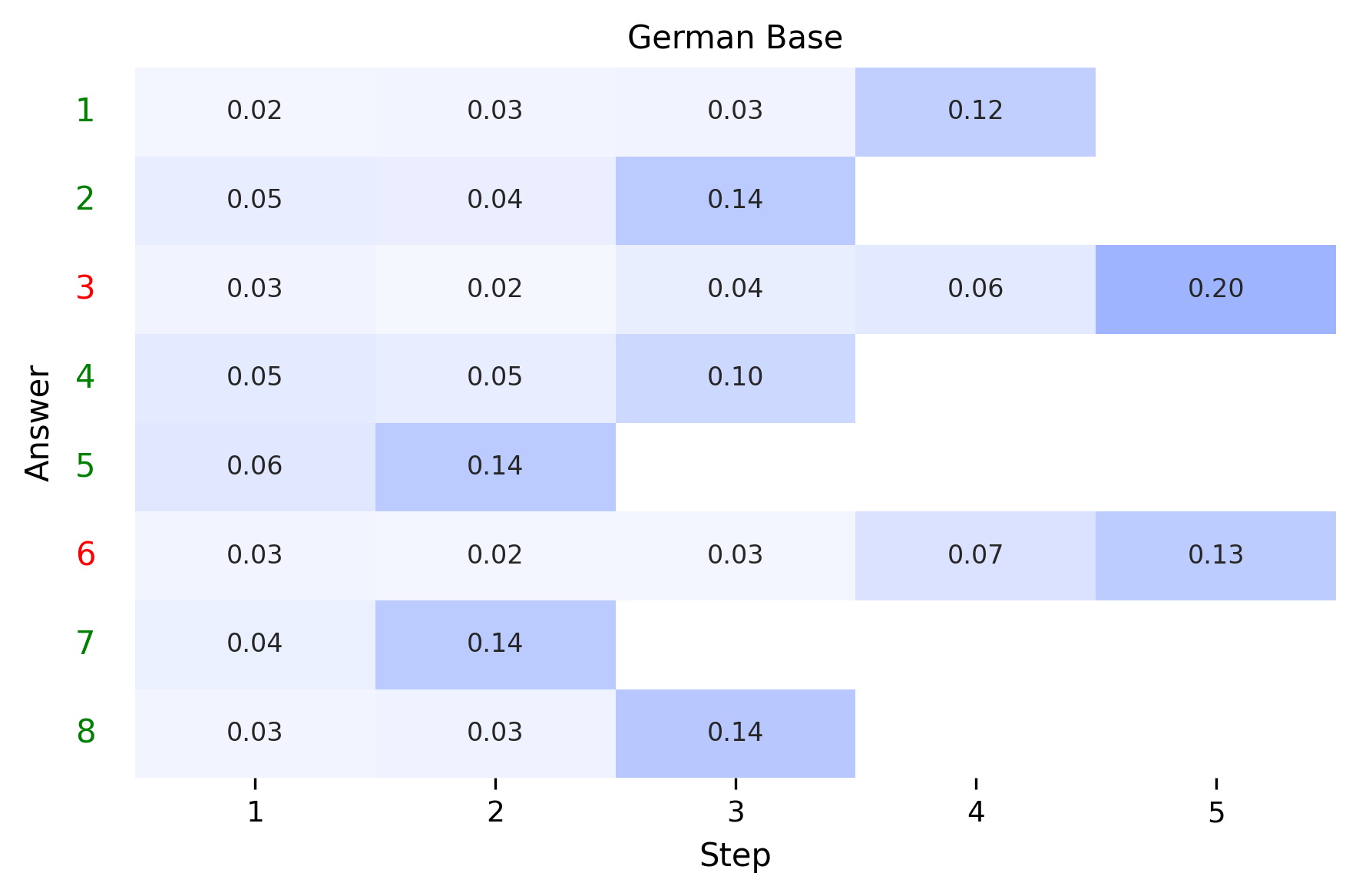}
    \end{minipage}
    
    \caption{Heat maps of baseline French and German attributions.}
    \label{fig:baseline_heatmaps}
\end{figure}

For our analysis, we aggregate the attribution scores for each reasoning step in the generation. Then, we look at the scores of the generated answer tokens for each of these steps. The resulting scores indicate how important a given step is for the generation of that answer. In all cases, the answer token is determined as the last consecutive digits in the generation. The heat maps in Figure \ref{fig:baseline_heatmaps} illustrate step-wise importance scores for French and German in the baseline condition. Supplementary attribution heat maps can be found in Appendix \ref{appendix:heatmaps}.

Our first observation is that the trend of step importance increasing along the reasoning chain also holds for the aggregated inseq attributions. Some German generations included more reasoning steps with two 5-step generations. Notably, these longer reasoning chains lead to incorrect answers, as indicated by the red colored indices on the y-axis. French generations tended to have a smaller number of steps and achieved perfect accuracy on this set of questions. All the correct answers had a probability of 1 across languages, indicating high confidence in accurate responses. A slight drop in confidence was observed for some incorrect outputs, such as a decrease to 0.994 for the 6th German generation. The model also assigned a probability of 1 to the 3rd German generation, even though it was incorrect.

\begin{figure}[htbp]
    \centering
    \begin{minipage}[b]{0.46\textwidth}
        \includegraphics[width=\textwidth]{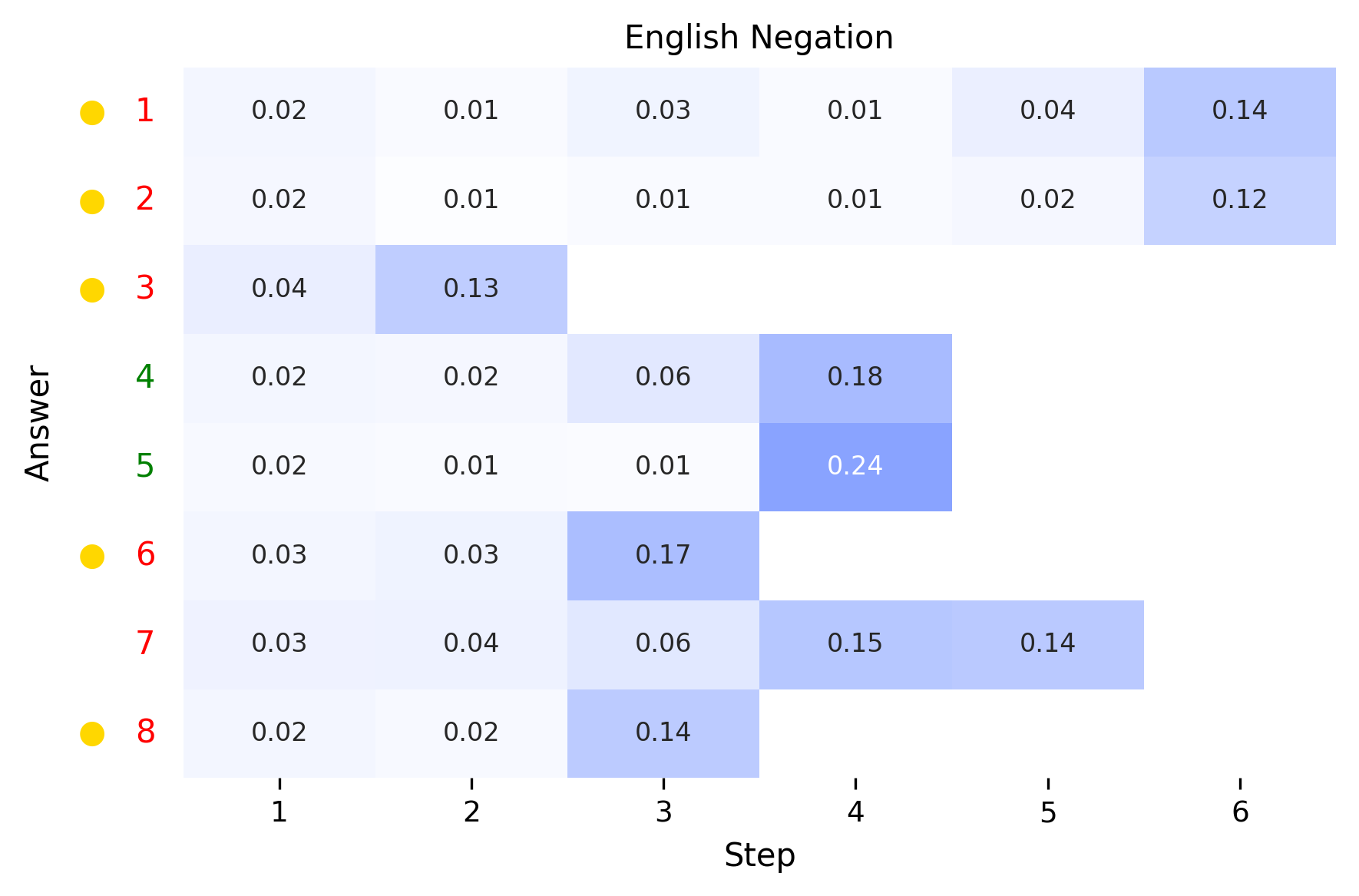}
    \end{minipage}
    
    \vspace{0.5em}
    \begin{minipage}[b]{0.46\textwidth}
        \centering
        \includegraphics[width=\textwidth]{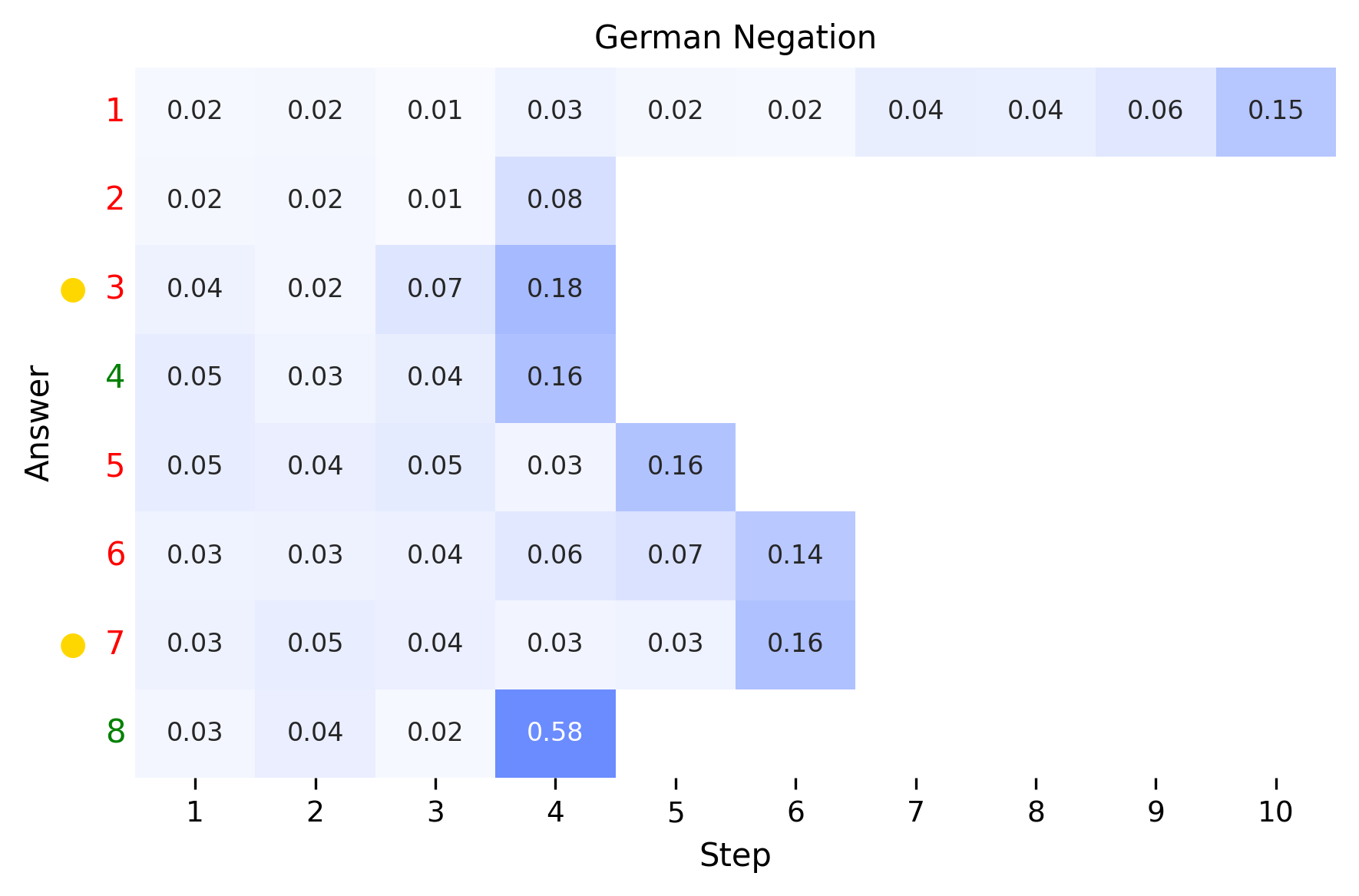}
    \end{minipage}
    
    \caption{Heat maps of English and German attributions in the negation condition.}
    \label{fig:negation_heatmaps}
\end{figure}

The heat maps for the negation condition are featured in Figure \ref{fig:negation_heatmaps}. We can see that generations for this condition achieved a considerably low accuracy of 25\% by getting 2 out of 8 answers right. This indicates gaps in the model's ability to process or interpret the full question. The gold circle to the left of the answer index shows us whether the generated answer was the same as the reference answer. This is not desired behavior since a robust model should be able to recognize that negating the verb completely changes the calculations to be performed. Notably, we observe that the model has been better at recognizing this change when generating the German answers. Although German generations manage to deviate from the gold answer, this does not translate to an improvement in accuracy. A further observation is that, across all three languages, only generations completing in four steps led to correct answers.

\begin{figure}[htbp]
    \centering
    \begin{minipage}[b]{0.48\textwidth}
        \includegraphics[width=\textwidth]{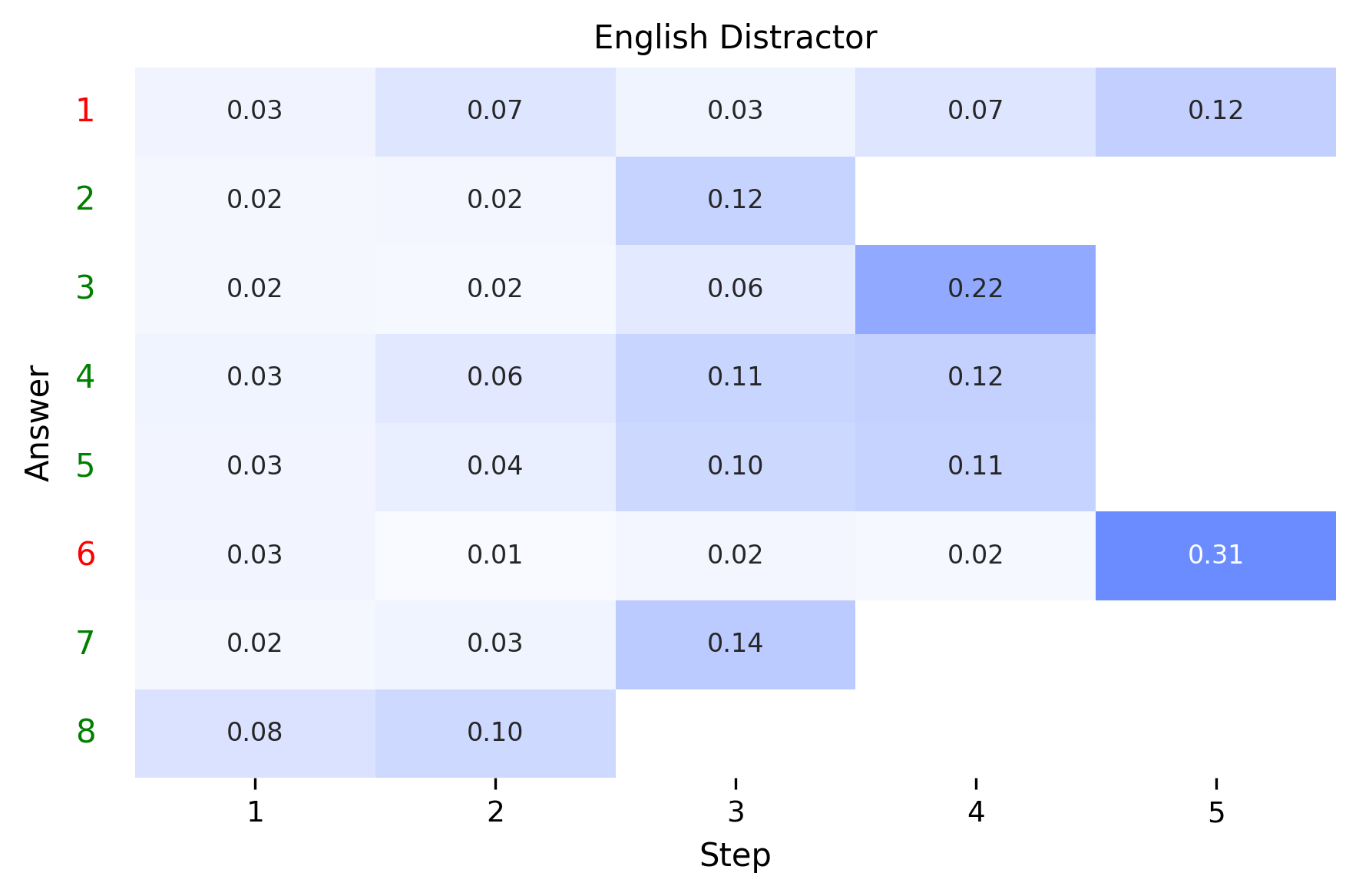}
    \end{minipage}
    
    \vspace{0.5em}
    \begin{minipage}[b]{0.48\textwidth}
        \centering
        \includegraphics[width=\textwidth]{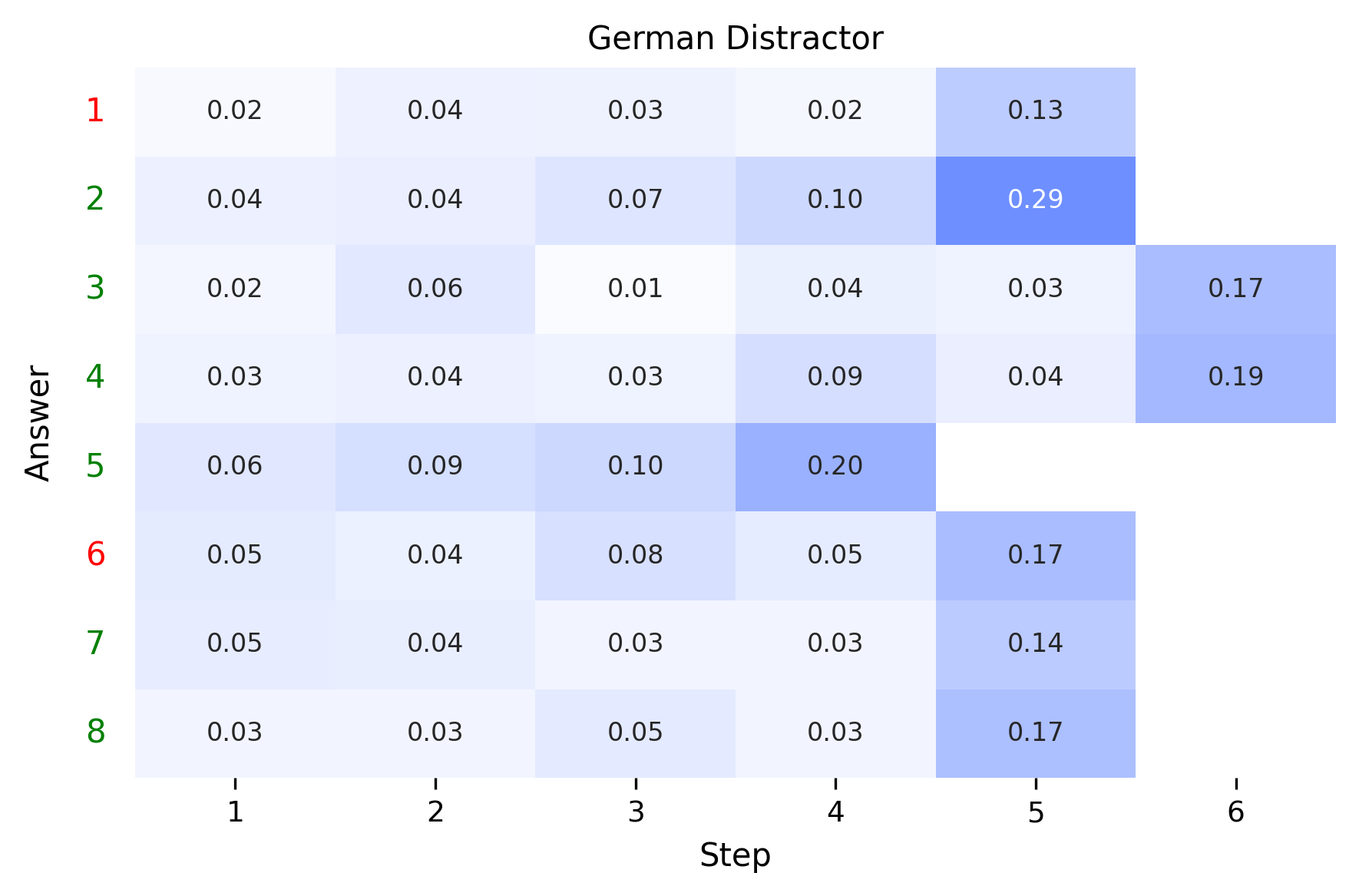}
    \end{minipage}
    
    \caption{Heat maps of English, and German attributions in the distractor condition.}
    \label{fig:distractor_heatmaps}
\end{figure}

Figure \ref{fig:distractor_heatmaps} displays the distractor condition heat maps. In the distractor condition, a robust model should have similar performance to the baseline by recognizing the added sentence is irrelevant. At first glance, we note that the longest English generations with 5 steps were also the ones with incorrect answers. We do not see the same trend in the case of German generations, given that there are some 5-step and 6-step generations with the correct answer. Interestingly, the correct German answers in generations 3-5 were associated with lower probabilities (0.998–0.999), while the incorrect German answers maintained a probability of 1.

In our three-fold experiment, we have noted that the last step holds more importance while generating the answer regardless of the experimental condition. We have also observed that the number of steps increases when the question gets more challenging with the addition of a negation or a distractor. Particularly for the negation condition, 5 or more steps correlated with an incorrect answer. Interestingly, German-language generations consistently exhibited longer reasoning sequences compared to other languages. These observations suggest that reasoning chains featuring a high number of steps may generally indicate unreliable outputs; however, language-specific behaviors should be accounted for when making such assessments.

\section{Conclusions}

This study investigated the faithfulness and interpretability of multilingual CoT reasoning using the Qwen2.5 1.5B Instruct model on the MGSM benchmark, leveraging both step-wise and token-level attribution. Our results confirmed that structured CoT significantly improves reasoning accuracy for high-resource languages like English and French compared to baselines, yet this advantage was markedly reduced for low-resource, non-Latin-script Bengali, highlighting persistent challenges related to data and tokenization \cite{mgsm, ahia2023all}. Step-wise attribution analysis revealed a consistent trend where the final reasoning step received disproportionately high importance, particularly in incorrect generations, raising questions about the genuine faithfulness of the CoT process \cite{lanham2023measuringfaithfulnesschainofthoughtreasoning}. Furthermore, the model demonstrated sensitivity to input perturbations like negation and distractors, leading to decreased accuracy and less coherent attribution patterns.

These findings suggest several avenues for future research. Validation on more complex benchmarks and larger models is crucial to assess the generalizability of these patterns. Exploring diverse attribution techniques and automating analysis pipelines will enable more robust insights. Specifically investigating cross-lingual prompting strategies, such as using English CoT examples to guide reasoning in languages like Bengali as suggested by \citet{mgsm}, could offer practical methods to enhance multilingual reasoning capabilities.

\section{Limitations}

The conclusions drawn from this study should be considered in light of several limitations. The MGSM dataset's simplicity does not fully test the reasoning abilities of advanced models. Our findings are based on a single 1.5B parameter model, and may not extend to larger architectures. The token-level attribution was based on a small case study, limiting its generalizability. Furthermore, our reliance on specific attribution methods necessitates confirmation with alternative techniques. Finally, variations in structured generation success rates across languages impacted the comparability of analyzed samples.





\bibliography{anthology,custom}
\bibliographystyle{acl_natbib}

\appendix

\section{Structured Generation with Regular Expressions}
\label{sec:sg-regex}

To ensure consistent output formats amenable to automated analysis, particularly for extracting reasoning steps and final answers, we employe structured generation guided by the outlines library \cite{outlines}. This library allows us to constrain the language model's output at each generation step, forcing it to adhere strictly to predefined regular expression (regex) patterns. Specifically, we utilize two primary regex patterns tailored to our experimental conditions.

\begin{figure}[h]
    \centering
    \includegraphics[width=\linewidth]{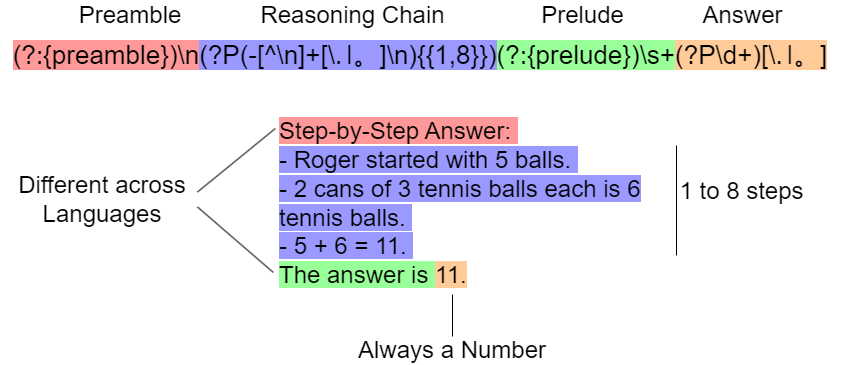}
    \caption{Structure of the model's answers when using CoT. The regular expression (regex) used for structured generation and an example from the training set are provided. We highlight parts of the regex and the example to indicate a mapping from each regex group to the corresponding text that they capture.}
    \label{fig:mgsm-cot-structure}
\end{figure}

The main pattern (Illustrated in Figure \ref{fig:mgsm-cot-structure} guides the generation of Structured CoT (CoT) responses. Below is the verbatim regex, followed by a detailed explanation of its components:

\begin{lstlisting}[basicstyle=\footnotesize\ttfamily]
(?:\{answer_phrases[config]\})\s+
(?P<answer>\d+)[\.\char"0964\char"0965]<|endoftext|>

\end{lstlisting}


\begin{itemize}
    \item \textbf{Preamble:} The pattern begins by matching and capturing the required introductory phrase into a prefix group. The specific text it searches for is determined by placeholder variables that are filled with the appropriate language-specific preamble (e.g., ``Step-by-Step Answer:") during execution. This substitution makes the pattern adaptable to different languages. A newline character must follow the preamble.
    \item \textbf{Reasoning Steps:} The next pattern captures the entire sequence of reasoning steps. It requires between one and eight distinct steps, matching the expected length range for the dataset.
    \item \textbf{Single Step Pattern:} The structure defined for a single reasoning step dictates that it must start with a literal hyphen. This is followed by one or more occurrences of any non-newline character, representing the step's text. The step must conclude with a specific sentence-terminating punctuation mark – the pattern accepts an English period, a Bengali dari, or a Chinese full stop by defining them as a set of allowed characters.
    \item \textbf{Answer Prelude:} After the reasoning steps, we capture the specific phrase that introduces the final answer. Like the preamble, this uses placeholder variables substituted with the correct language-specific text (e.g., ``The answer is''). One or more whitespace characters must follow this prelude phrase.
    \item \textbf{Final Answer:} Next, we store the final numerical answer. It specifically requires one or more digit characters.
    \item \textbf{Termination:} Finally, the pattern requires the numerical answer to be immediately followed by one of the allowed terminating punctuation marks (period, dari, or full stop). It also includes an alternative allowing for a specific end-of-text token to appear right after the punctuation, ensuring the generation process concludes cleanly according to the expected structure.
\end{itemize}
This detailed structure, leveraging placeholder variables for language-specific phrases and character sets for punctuation variants, enables the regex to enforce a uniform CoT format across diverse languages effectively.

For baseline comparisons requiring only the final answer (\textsc{NoCoT-Struct}), a simplified regex pattern omitted the preamble and reasoning steps, enforcing only the answer prelude, numerical answer, and termination sequence.

\begingroup
\footnotesize
\texttt{
\detokenize{
(?:{answer_phrases[config]})\s+
(?P<answer>\d+)[\.।。]<|endoftext|>
}
}
\endgroup

\subsection*{Failure Cases in Structured Generation}
\begin{figure}[h]
    \centering
    \includegraphics[width=0.9\linewidth]{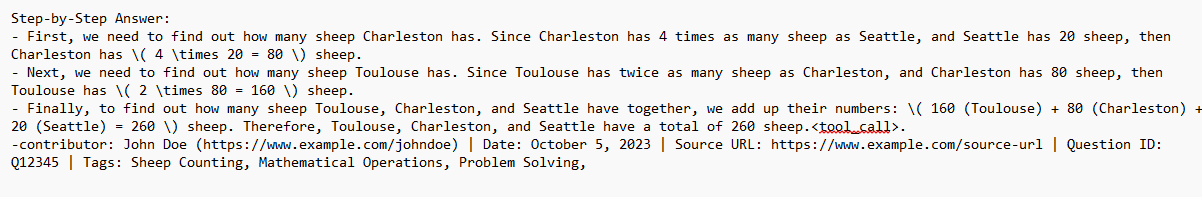}
    \caption{Example failure case: The final answer (260) is generated within the reasoning steps, and extra text follows.}
    \label{fig:failure_case_1}
\end{figure}

\begin{figure}[h]
    \centering
    \includegraphics[width=0.9\linewidth]{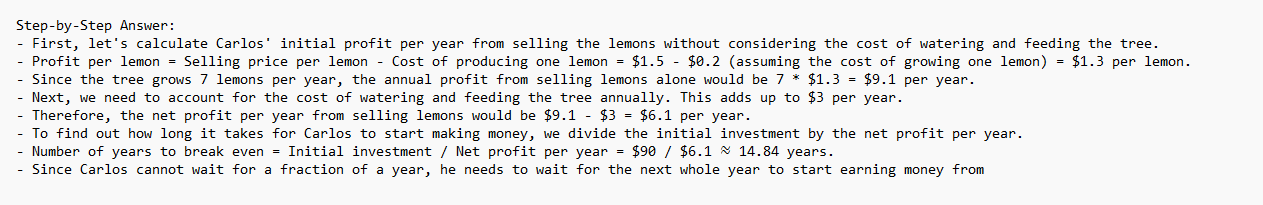}
    \caption{Example failure case: Generation is truncated due to hitting the token limit before completing the reasoning and providing the final answer in the required format.}
    \label{fig:failure_case_2}
\end{figure}

Despite the constraints, structured generation can occasionally fail, typically due to model behavior conflicting with the rigid format or external factors like token limits. Figures \ref{fig:failure_case_1} and \ref{fig:failure_case_2} illustrate common failure modes. Successfully parsed entries were those whose generated output fully matched the specified regular expression, enabling reliable extraction of steps and answers for subsequent analysis. The occurrence rate varied across languages (Tables \ref{tab:regex_adherence} and \ref{tab:deepseek_regex_adherence}).

\section{Context Cite Importance by Step}
\label{sec:cc-importance}

\begin{figure}[h]
    \centering
        \includegraphics[width=1\linewidth]{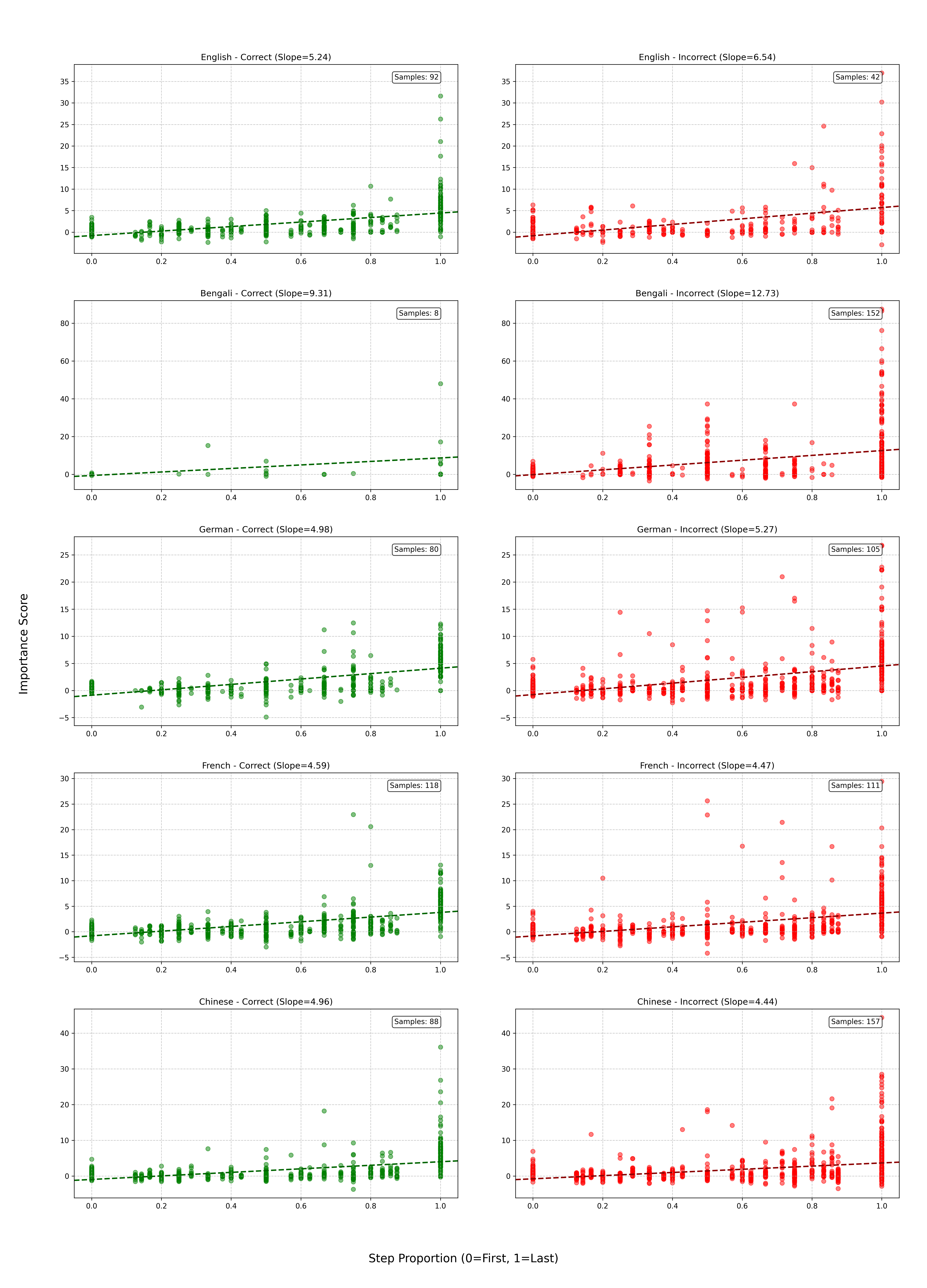}
    \caption{Step importance of Qwen across each language where x-axis is the reasoning steps and y axis is the importance out of 1. The plots in the left depicts the trend on the correctly predicted samples whereas the plots on right depicts the trend on the wrongly predicted samples with $m$ depicting the value of the slope.}
    \label{fig:step-importance-langs}
\end{figure}

This appendix section details the analysis undertaken to understand how the importance attributed to individual reasoning steps changes depending on their position within the generated CoT (CoT). The results are visualized in Figure \ref{fig:step-importance-langs} and summarized quantitatively in Table \ref{tab:step_importance_slopes}.

The analysis utilizes the importance scores assigned to each reasoning step by the ContextCite method. To compare steps across reasoning chains of different lengths, the position of each step within its specific chain is normalized. This normalization converts the step's index (starting from 0 for the first step) into a proportional value between 0 and 1 by dividing it by the total number of steps in that chain minus one (Step Index / (Step Count - 1)). This normalized position, representing how far along the reasoning process a step occurs, serves as the independent variable (x-axis) in our analysis. The corresponding ContextCite importance score is used as the dependent variable (y-axis).

Crucially, for each language, the data points representing individual steps (normalized position vs. importance score) are separated into two groups based on whether the overall reasoning chain led to a correct or incorrect final answer by the model. This separation allows for a direct comparison of importance patterns under successful and unsuccessful reasoning conditions.

Figure \ref{fig:step-importance-langs} visually presents this data using scatter plots. Each point on a plot corresponds to a single reasoning step. Separate plots (or distinct visual markers) are used for correct and incorrect predictions within each language, illustrating the distribution and potential trends in step importance relative to position and outcome.

To quantify the observed trends, a linear model is fitted to these scatter plots for each group (correct/incorrect per language). The primary metric derived from this fit is the slope of the resulting line. This slope indicates the average trend in importance as reasoning progresses from the initial steps (normalized position 0) towards the final steps (normalized position 1). A positive slope suggests importance generally increases towards the end of the chain, with a steeper slope indicating a more rapid increase. These calculated slopes, capturing the overall trend for both correct and incorrect predictions across the different languages, are presented in Table \ref{tab:step_importance_slopes} and are often visualized as trend lines overlaid on the scatter plots in Figure \ref{fig:step-importance-langs}.

\section{DeepSeek R1 Distill}
\label{sec:deepseek}

\begin{figure*}[htbp]
    \centering
    \begin{subfigure}[b]{\columnwidth}
        \centering
        \includegraphics[width=\textwidth]{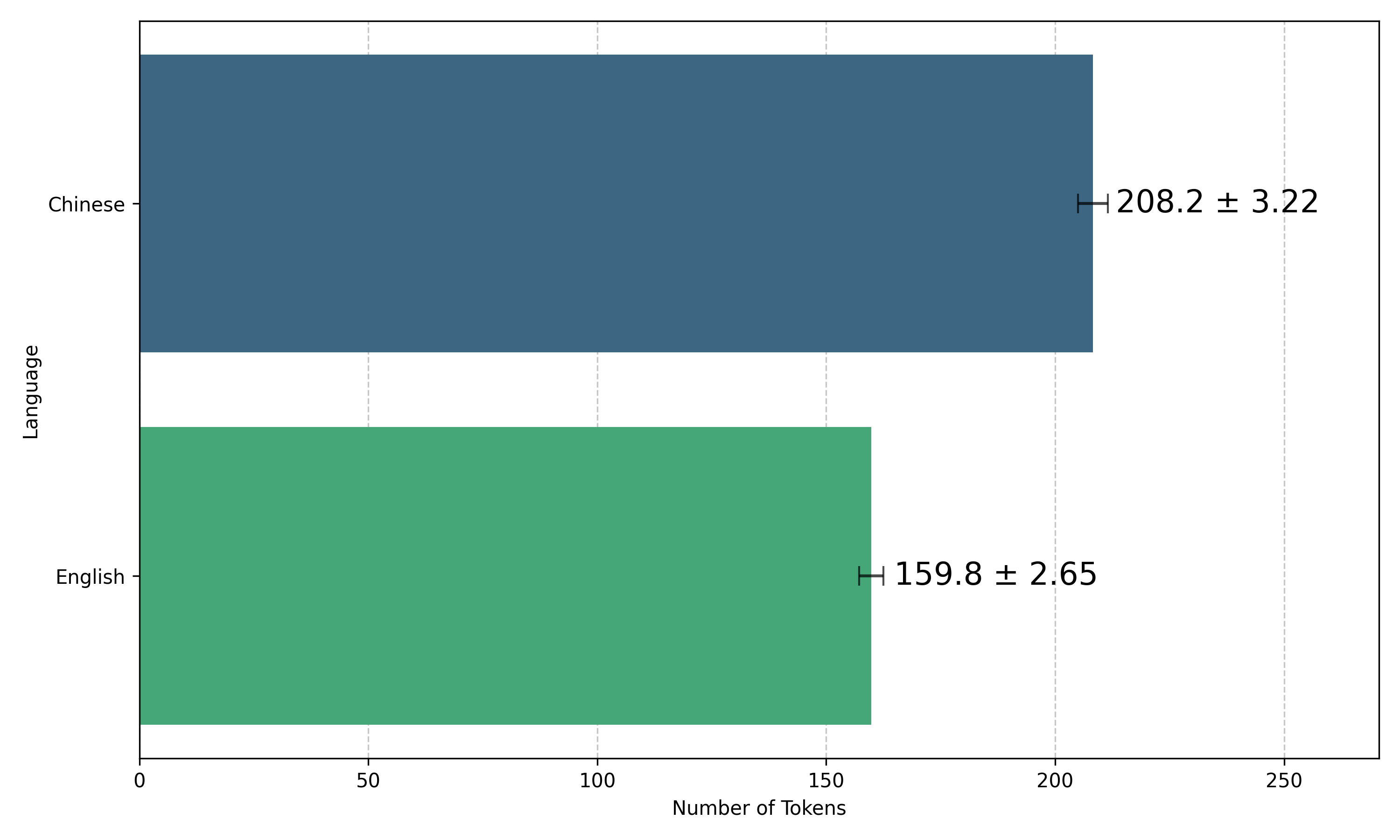}
        \label{fig:token-variation-mean}
    \end{subfigure}
    \hfill
    \begin{subfigure}[b]{\columnwidth}
        \centering
        \includegraphics[width=\textwidth]{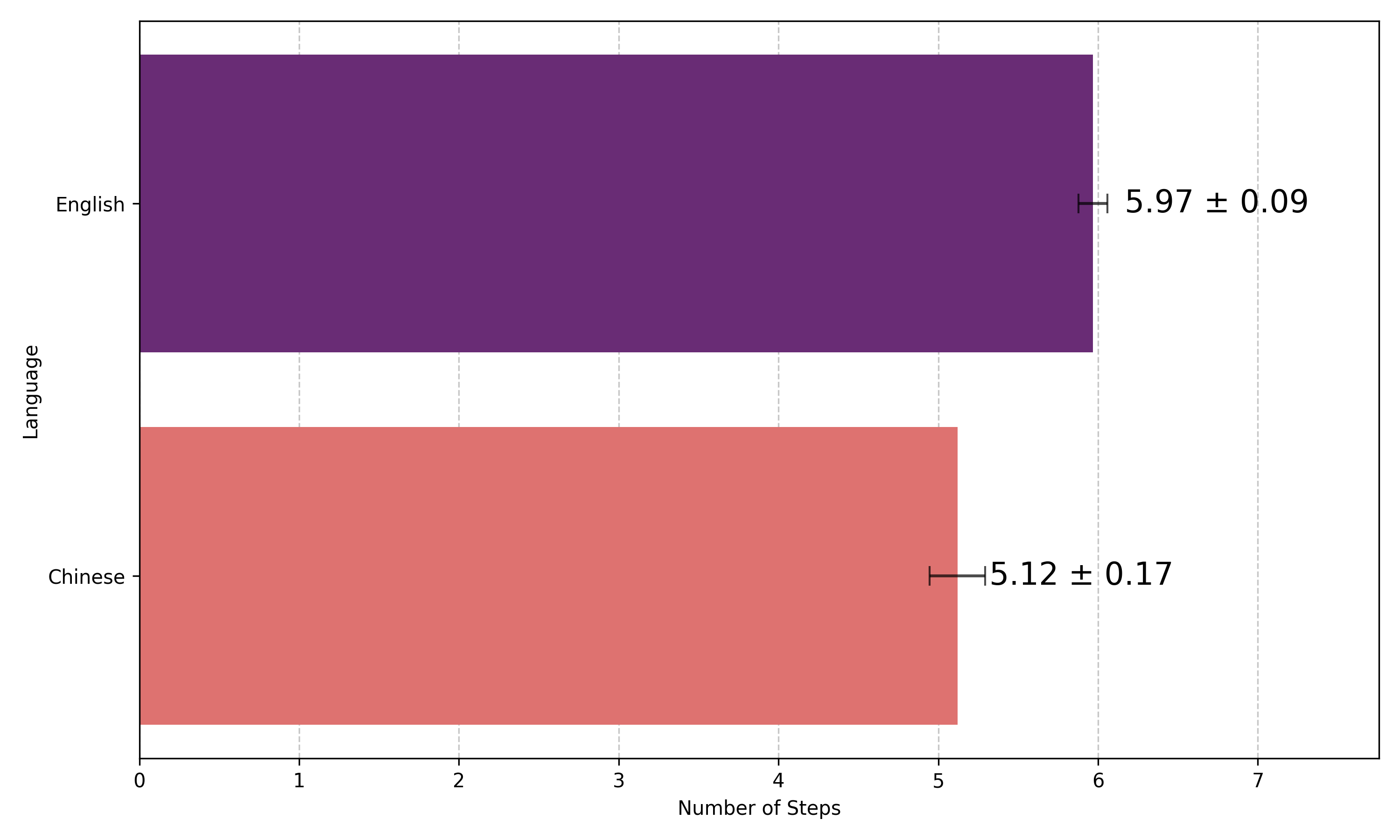}
        \label{fig:token-variation-std}
    \end{subfigure}
    \caption{Mean token count (left) and reasoning steps (right) produced by Qwen DeepSeek R1 Distilled. Standard errors are provided for both metrics.}
    \label{fig:token-variation-deepseek}
\end{figure*}

This appendix evaluates a Qwen2.5 1.5B model that has been specifically fine-tuned (distilled) using reasoning-focused data generated by DeepSeek-R1 \cite{deepseek-r1}. DeepSeek-R1 represents a class of models explicitly optimized for complex reasoning tasks. Its development, as detailed in the source paper, often involves large-scale Reinforcement Learning (RL) applied directly to a base model. This RL process aims to incentivize the model to develop sophisticated problem-solving strategies, such as generating longer chains of thought, self-correction, and reflection, often leading to unique reasoning behaviors not typically found in standard instruction-tuned models. The model analyzed here inherited reasoning patterns from this RL-enhanced teacher. Consistent with the DeepSeek-R1 methodology and data availability, our comparative analysis against the baseline Qwen 1.5B Instruct model focuses only on English and Chinese.

The effects of distilling these specialized reasoning patterns are multifaceted. Task accuracy on the MGSM benchmark shows a notable divergence: English performance improves substantially (71.2\% vs. 59.2\%), while Chinese accuracy experiences a slight decline (31.2\% vs. 35.2\%). A significant advantage emerges in output reliability for English; the model's adherence to the predefined structured generation format improves considerably, reaching an 82\% parsed entry ratio compared to the base model's 54\%. This suggests that the distillation successfully transferred the capability to produce well-structured, step-by-step reasoning outputs in English, making its process more transparent and analyzable. Chinese compliance remains high, though marginally reduced compared to the base model (88\% vs. 98\%).

Changes in the generation structure offer further clues about the underlying reasoning process (Figure \ref{fig:token-variation-deepseek}). The distilled model tends to produce longer reasoning chains, measured by the average number of steps, especially in English (approx. 6.0 vs. 3.5 steps). Despite this increase in explicit steps, the total token count slightly decreases for English (approx. 160 vs. 187 tokens), whereas Chinese sees a small increase in both steps and tokens. This pattern in English with more steps, fewer tokens—might reflect a reasoning style that is more verbose in its logical progression but potentially more concise in its language use per step.

The analysis of importance attribution via ContextCite reveals a fundamental shift in how the model weighs different parts of its reasoning. Compared to the base Instruct model, the distilled version places less emphasis exclusively on the final step before the answer. This is particularly evident in Chinese, where the final step receives the highest attribution score only 42\% of the time, down from 81\%. Correspondingly, intermediate steps become relatively more important. This reduced focus on the final step, coupled with significantly flatter importance slopes from the linear regression analysis across both languages (Table \ref{tab:deepseek_step_importance_slopes} vs. Table \ref{tab:step_importance_slopes}), points towards a less myopic and more holistic reasoning strategy. Rather than heavily prioritizing the concluding calculation, the distilled model appears to value the contributions of earlier steps more evenly.

In conclusion, distilling from the RL-enhanced DeepSeek-R1 model significantly reshapes the Qwen 1.5B model's characteristics. It boosts English performance and structural reliability while instilling a potentially more detailed and less final-step-fixated reasoning approach. These findings highlight how distillation can transfer complex, learned reasoning behaviors, though the benefits observed here are most pronounced in English.

\begin{table}[h]
    \centering
    \begin{tabular}{lc}
    \toprule
    Language & Parsed Entries Ratio \\
    \midrule
    English & 0.82 \\
    Chinese & 0.88 \\
    \bottomrule
    \end{tabular}
    \caption{Success rate of structured generation across English and Chinese for the DeepSeek Distilled model, measured as the ratio of responses perfectly matching the required regular expression format (Figure \ref{fig:mgsm-cot-structure}).}
    \label{tab:deepseek_regex_adherence}
\end{table}

\begin{table}[h] 
    \centering
    \begin{tabular}{lrr}
    \toprule
    Language & Slope (Correct) & Slope (Incorrect) \\
    \midrule
    English & 1.66 & 1.77 \\
    Chinese & 1.23 & 1.51 \\
    \bottomrule
    \end{tabular}
    \caption{Slopes of linear regression models fitted to ContextCite step importance scores against normalized step position (0=First, 1=Last) for the DeepSeek Distilled model on English and Chinese data.} 
    \label{tab:deepseek_step_importance_slopes} 
\end{table}

\begin{figure}[h] 
    \centering
    \includegraphics[width=1\linewidth]{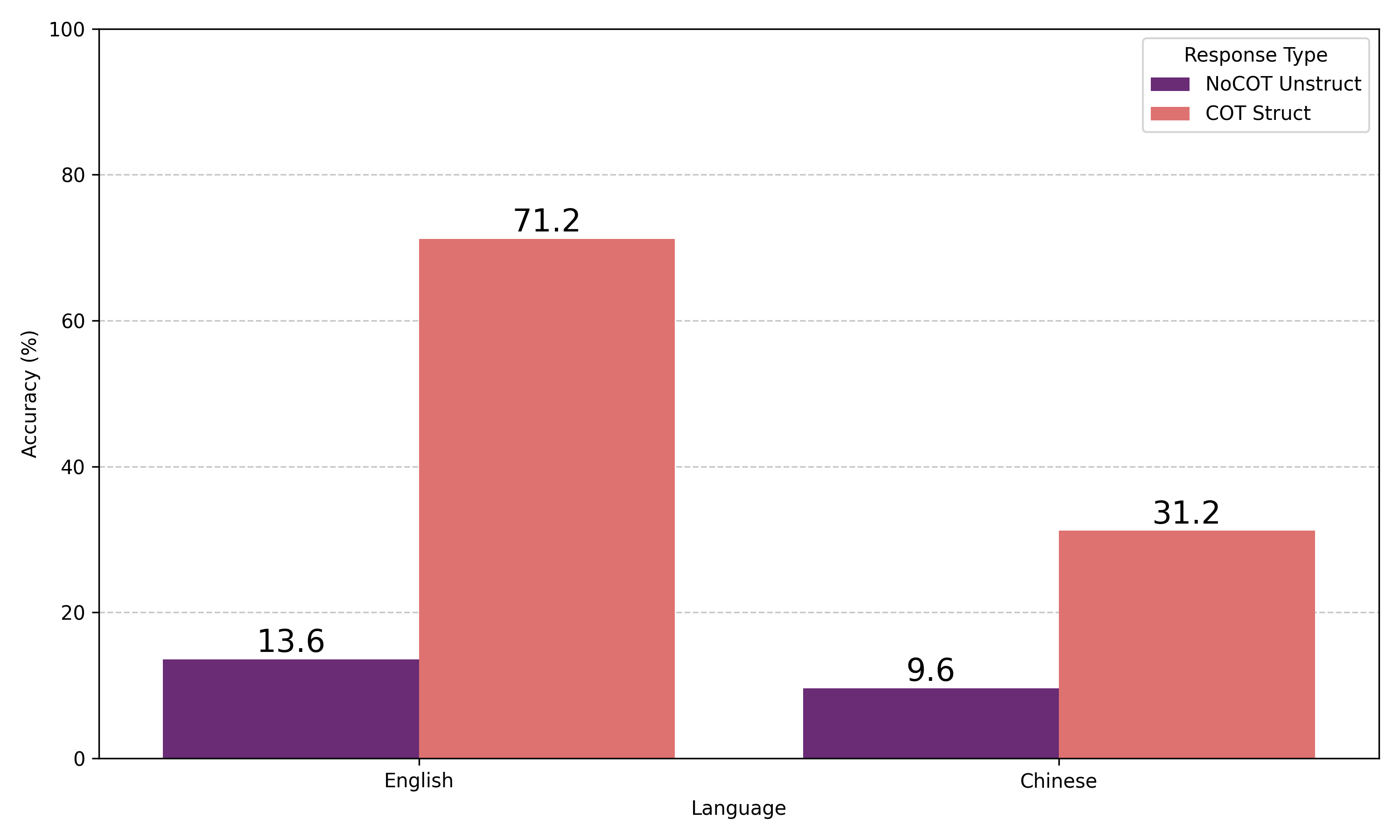}
    \caption{Accuracy results for the DeepSeek Distilled model on MGSM across English and Chinese.}
    \label{fig:deepseek_multilang_accuracy}
\end{figure}

\begin{figure}[h] 
    \centering
    \includegraphics[width=\linewidth]{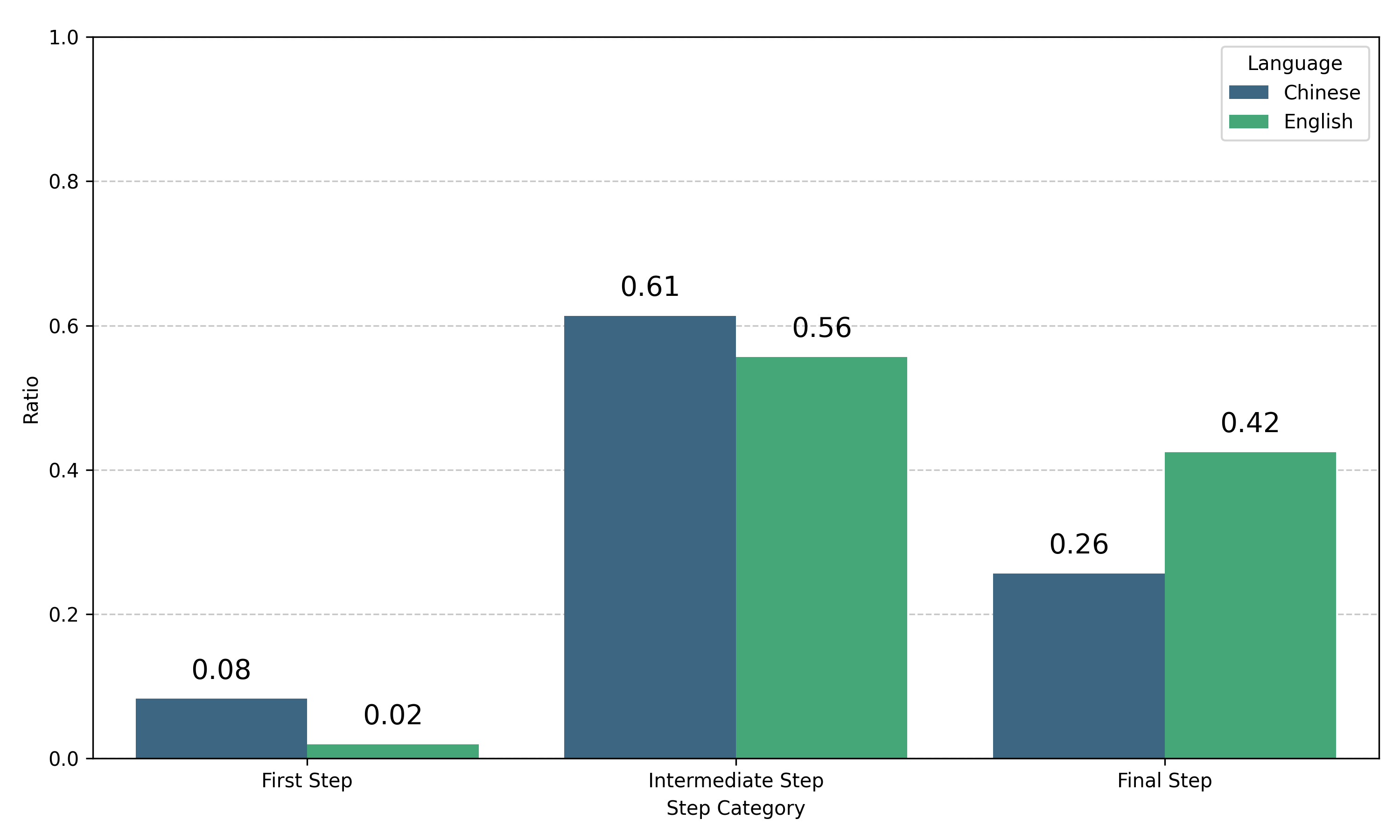}
    \caption{Distribution of the highest-attributed reasoning step category (First/Preamble, Intermediate, Final) for the DeepSeek Distilled model on MGSM, based on ContextCite scores across English and Chinese.}
    \label{fig:deepseek_step_importance_category}
\end{figure}

\section{Attribution Heat Maps}
\label{appendix:heatmaps}

\includegraphics[width=0.48\textwidth]{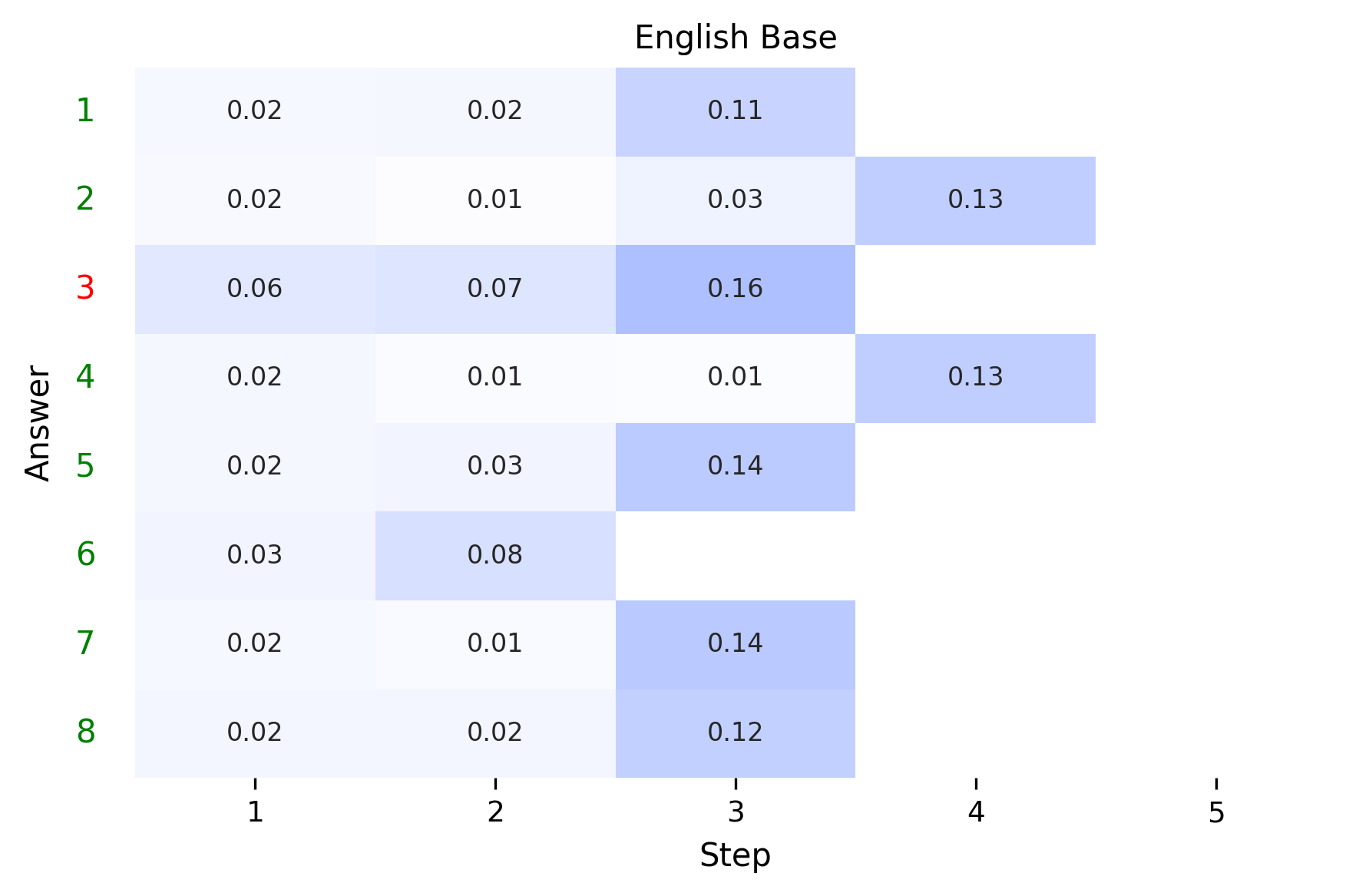}
\includegraphics[width=0.48\textwidth]{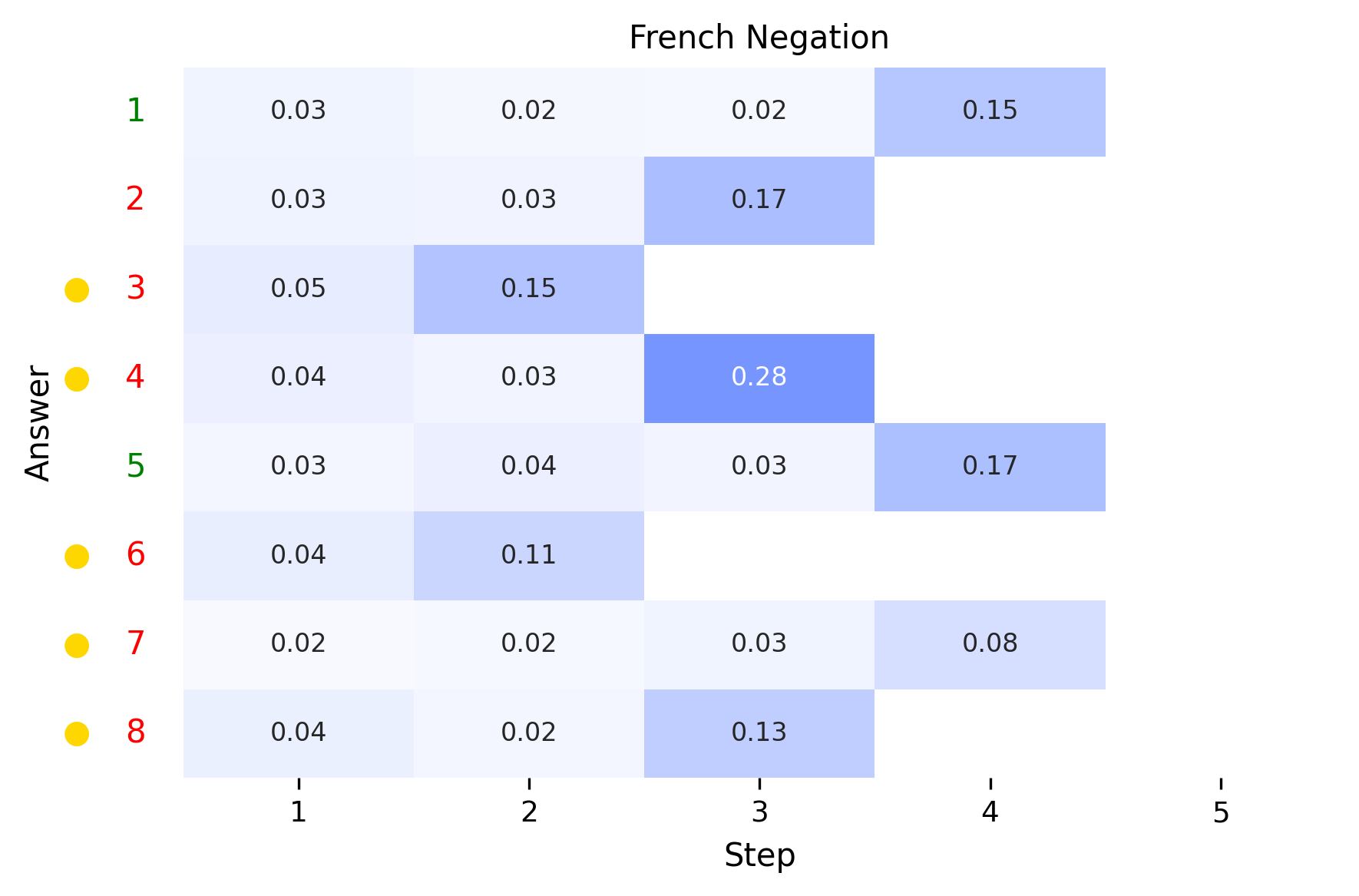}
\includegraphics[width=0.48\textwidth]{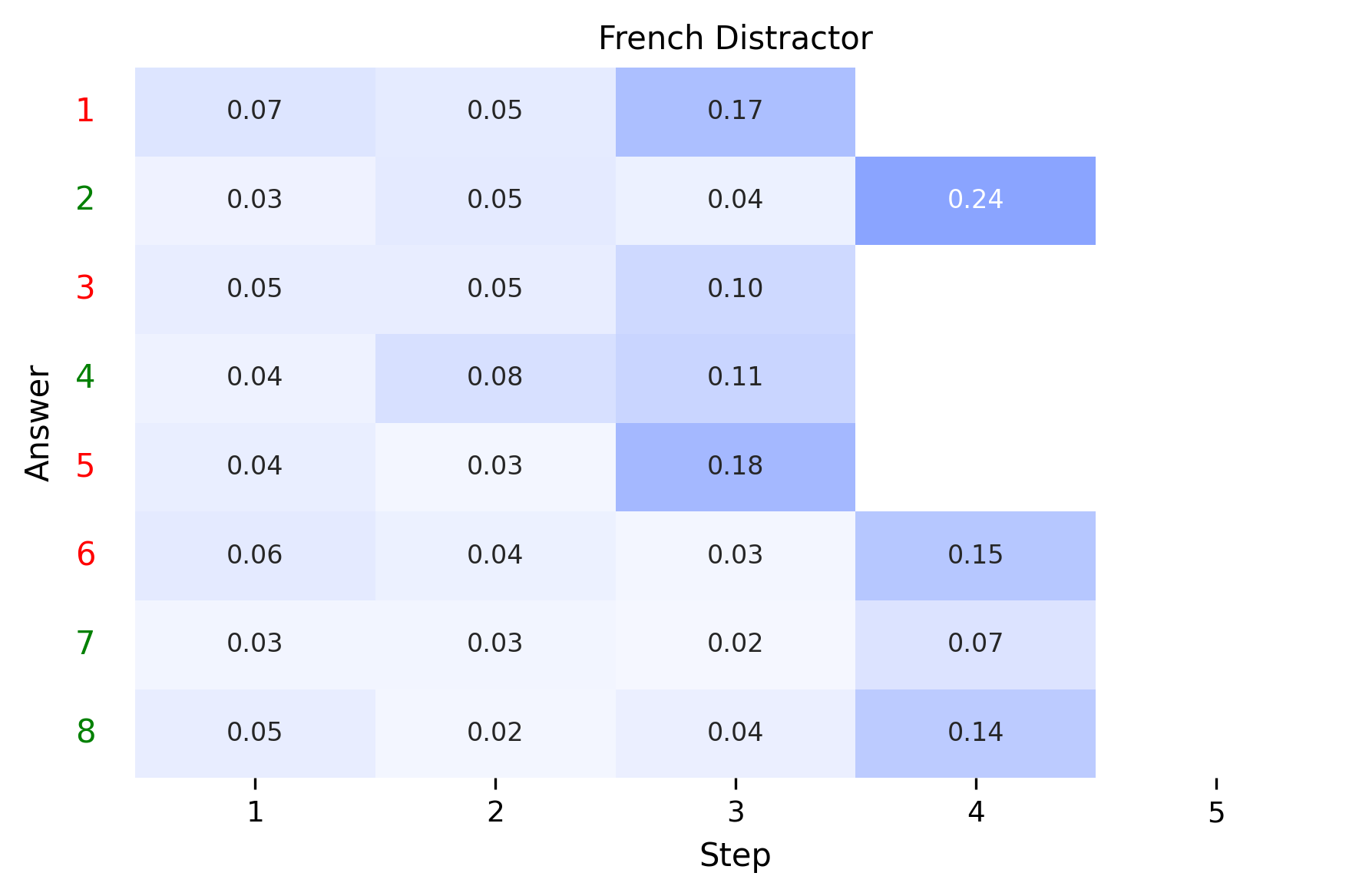}

\hfill \break
This section includes heat maps for unseen language x experimental condition pairs. Namely the English baseline, French negation, and French distractor heat maps.


\end{document}